\pdfoutput=1
\documentclass{article} %
\usepackage[dvipsnames]{xcolor}
\usepackage{iclr2023_conference,times}

\usepackage{amsmath,amsfonts,bm}

\def\eqref#1{equation~\ref{#1}}

\def\1{\bm{1}}

\def\vf{{\bm{f}}}

\def\vm{{\bm{m}}}

\def\vx{{\bm{x}}}

\def\vz{{\bm{z}}}

\DeclareMathAlphabet{\mathsfit}{\encodingdefault}{\sfdefault}{m}{sl}
\SetMathAlphabet{\mathsfit}{bold}{\encodingdefault}{\sfdefault}{bx}{n}

\newcommand{\dB}{\ensuremath{~\mathrm{dB}}}
\newcommand{\FID}{\ensuremath{~\mathrm{FID}}}

\usepackage{url}
\usepackage{graphicx}
\usepackage{subcaption}
\colorlet{linkcolor}{violet}
\colorlet{citecolor}{RedOrange}  %
\colorlet{urlcolor}{Aquamarine}

\usepackage{hyperref}
    \newcommand\myshade{85}
    \hypersetup{
      linkcolor  = linkcolor!\myshade!black,
      citecolor  = citecolor!\myshade!black,
      urlcolor   = urlcolor!\myshade!black,
      colorlinks = true,
    }

\usepackage[capitalise, nameinlink]{cleveref}
    \Crefname{table}{Tab.}{Tabs.}
    \Crefname{appendix}{App.}{Apps.}
    \Crefname{section}{Sec.}{Secs.}
    \Crefname{equation}{Eq.}{Eqs.}

\usepackage{array}
    \newcolumntype{C}{>{$}c<{$}}
    \newcolumntype{R}[2]{%
        >{\adjustbox{angle=#1,lap=\width-(#2)}\bgroup}%
        l%
        <{\egroup}%
        }
    \newcommand*\rottab{\multicolumn{1}{R{45}{1em}}}
\usepackage{multirow}
\usepackage{booktabs}
\usepackage[export]{adjustbox}
\usepackage{standalone}
\usepackage[normalem]{ulem}
\usepackage[inline]{enumitem}
\usepackage{xspace}
\usepackage{wrapfig}
\usepackage{pifont}
\newcommand{\cmark}{\text{\ding{51}}}
\newcommand{\xmark}{\text{\ding{55}}}

\usepackage{tikz}
\usetikzlibrary{positioning}
\usetikzlibrary{calc}
\usetikzlibrary{backgrounds}

\usepackage{microtype}

\newcommand{\functa}{{functa}\xspace}
\newcommand{\sfuncta}{{spatial functa}\xspace}

\title{Spatial Functa: Scaling Functa to ImageNet Classification and Generation}

\author{
	\hspace{-5mm}Matthias Bauer$^{*1}$ \quad Emilien Dupont$^{1}$ \quad Andy Brock$^{1}$ \quad Dan Rosenbaum$^{2}$ \quad Jonathan Schwarz$^{1}$\\
	\hspace{56mm} \textbf{Hyunjik Kim}$^{*1}$ \\ \\
	\hspace{42mm}$^{1}$DeepMind, $^{2}$University of Haifa
}

\footnotetext[1]{Equal Contribution. Correspondence to: \texttt{\{hyunjikk,msbauer\}@google.com}.}

\iclrfinalcopy %
\begin{document}

\maketitle

\begin{abstract}
Neural fields, also known as implicit neural representations, have emerged as a powerful means to represent complex signals of various modalities. Based on this \citet{functa} introduce a framework that views neural fields as data, termed \textit{functa}, and proposes to do deep learning directly on this dataset of neural fields. In this work, we show that the proposed framework faces limitations when scaling up to even moderately complex datasets such as CIFAR-10. We then propose \textit{spatial functa}, which overcome these limitations by using spatially arranged latent representations of neural fields, thereby allowing us to scale up the approach to ImageNet-1k at $256 \times 256$ resolution. We demonstrate competitive performance to Vision Transformers \citep{steiner2021train} on classification and Latent Diffusion \citep{rombach2022high} on image generation respectively.
\end{abstract}

\begin{figure}[h]
    \centering
    \includegraphics[width=\textwidth]{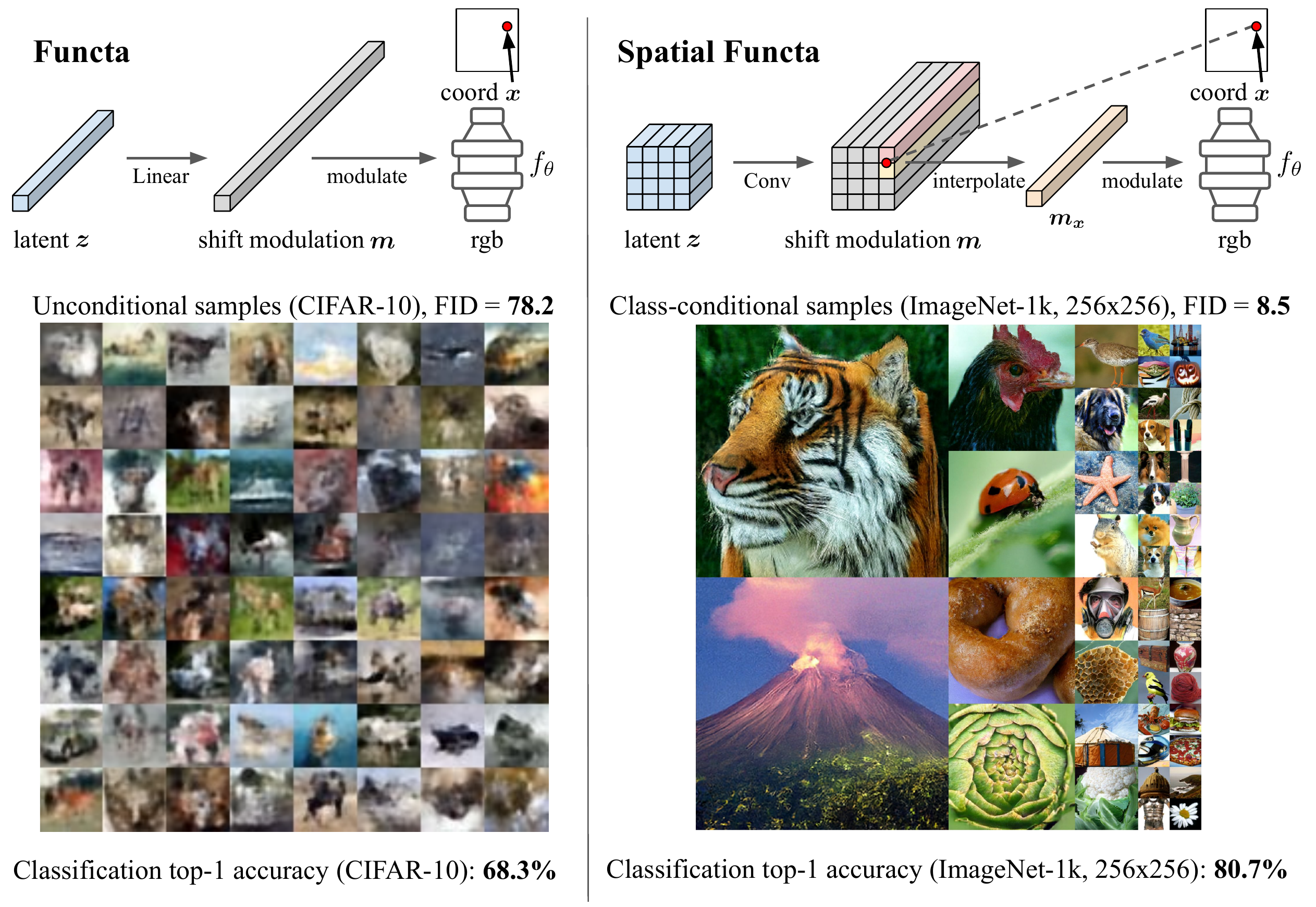}
    \caption{Comparing \textit{functa} \citep{functa} to our proposed \textit{spatial functa}. \textit{Top row:} neural field parameterizations.
    \textit{Bottom row:} curated %
    samples and classification accuracies on CIFAR-10 (functa) and ImageNet $256 \times 256$ (spatial functa), respectively, without pretraining on other datasets.}
    \label{fig:figure_1}
\end{figure}

\section{Introduction}
\label{sec:intro}
\textit{Neural fields}, also known as \textit{implicit neural representations} (INRs), are coordinate-based neural networks that represent a field, such as a 3D scene, by mapping 3D coordinates to colour and density values in 3D space.
Neural fields have recently gained significant traction in computer vision as a method to represent signals such as images \citep{ha2016generating}, 3D shapes/scenes \citep{park2019deepsdf, mescheder2019occupancy, mildenhall2020nerf}, video \citep{chen2021nerv}, audio \citep{sitzmann2020implicit}, medical images \citep{shen2022nerp} and climate data \citep{dupont2021generative}.
Building on the fact that neural fields can represent signals in such a wide range of modalities, \cite{functa} recently proposed a framework termed \textit{functa} for doing deep learning directly on these field representations, rather than the classical approach of processing array representations such as pixels. They show promising results on generation, inference and classification across a wide range of modalities. They cover images, voxels, climate data and 3D scenes, but on fairly small scale and simple datasets e.g. CelebA-HQ $64\times64$ \citep{karras2018progressive}, ShapeNet \citep{chang2015shapenet}.

In this work, we propose a method for scaling up functa to larger and more complex datasets. Concretely, we make the following contributions:
\begin{itemize}[topsep=-2pt,itemsep=4pt,partopsep=0pt, parsep=0pt, leftmargin=1em]
    \item After reproducing the reported functa results on CelebA-HQ, we apply the same methodology to downstream tasks on CIFAR-10, and show that -- surprisingly -- the results on classification and generation are far below expectations (see left of \cref{fig:figure_1} and \cref{tab:cifar10_functa}).
    \item We introduce \textit{spatial functa}, a spatially arranged representation of functa whose features capture local signal information in input space.
    \item We demonstrate that spatial functa overcome the limitations observed for CIFAR-10 classification/generation and can scale to much larger datasets, namely ImageNet-1k at $256 \times 256$ resolution. This is demonstrated by classification results that are competitive with ViTs \citep{steiner2021train} and image generation results competitive with Latent Diffusion \citep{rombach2022high}.
\end{itemize}

\section{Background on Neural Fields and Functa}
\label{sec:background}
\textbf{Neural fields} are functions $f_\theta$ mapping \textit{coordinates} $\vx$ (e.g.\ pixel locations) to \textit{features} $\vf$ (e.g.\ RGB values) parameterized by $\theta$. They are fit to a particular signal by minimizing the reconstruction loss between the output $\vf$ and the target signal value at all available coordinates $\vx$. For example, an image would correspond to a single $f_\theta$, usually an MLP with sinusoidal activation functions, i.e., a SIREN \citep{sitzmann2020implicit}. Neural fields are very general and can represent a wide range of the aforementioned modalities.

\textbf{Functa} \Citep{functa} is a framework that treats neural fields as data points, elements of a \textit{functaset}, on which we can perform deep learning tasks. See \cref{sec:app:background} for a more thorough introduction. The motivations for doing so include: 
\begin{enumerate*} [label=(\arabic*)]
    \item \textbf{better scaling} of data dimensionality compared to array representations, leading to more memory/compute efficient training of neural networks for downstream tasks;
    \item \textbf{moving away from a fixed resolution};
    \item allowing for \textbf{unified frameworks and architectures across multiple data modalities};
    \item dealing with \textbf{signals that are inherently difficult to discretize}, such as fields on non-Euclidean manifolds, e.g.,\ climate data.
\end{enumerate*} 
\Citet{functa} show promising results on relatively small scale datasets with limited complexity.
In this work we ask the question: \emph{does the approach scale to more complex / larger datasets? If not, how can we make it scale?}

\section{Limitations of naively scaling up Functa}
\label{sec:functa_limitations}

Following implementation details by \citet{functa}, we were able to reproduce reported results on CelebA-HQ ($64\times 64$) and subsequently applied \functa to CIFAR-10 ($32\times 32$). We discovered that, while the meta-learned functa latent representations could faithfully reconstruct the data, on downstream tasks they significantly underperformed pixel-based classification and generation.

More specifically, we were able to fit each CIFAR-10 image successfully with meta-learning, that is, we could reconstruct test set images to very high fidelity (38.1dB peak signal to noise ratio (PSNR)) on the test set with 1024 dimensional latents, see \cref{tab:cifar10_functa})\footnote{This PSNR value corresponds to a mean absolute error of $\approx 3$ units per RGB value (units in range $[0,255]$), which is hardly noticeable to the human eye, see~\protect\cref{fig:cifar10_functa}, \cref{sec:app:functa_limitations}}. Hence it is clear that the information necessary for downstream tasks is still present in the latent representation.
However, we found that both MLPs (with residual connections as successfully employed by \cite{functa} on simpler datasets) and Transformers struggle to make use of this information for downstream tasks: Starting from the model configurations and insights by \citet{functa}, we carried out a thorough hyperparameter search across architectures for both classification and generation using DDPM \citep{Ho2020}. We varied model sizes, optimizer configurations and explored standard regularization techniques such as data augmentation, model averaging and label smoothing (for classification) to push performance. However the best top-1 accuracy that we were able to achieve was 68.3\%, and the best FID for diffusion modeling was 78.2, c.f. samples in \cref{fig:figure_1} \textit{(left)}. See \cref{sec:app:functa_limitations} for details. 

\begin{wraptable}{r}{0.6\textwidth}
\vspace{-4mm}
\centering
\begin{tabular}{cccc}
\toprule
latent dim & Test PSNR & top-1 acc & FID (uncond) \\\midrule
256 & 27.6 dB & 66.7\% & \textbf{78.2} \\ 
512 & 31.9 dB & \textbf{68.3}\% & 96.1 \\ 
1024 & \textbf{38.1} dB & 66.7\% & 134.8 \\
\bottomrule
\end{tabular}
\caption{CIFAR-10 functa results}
\vspace{-6mm}
\label{tab:cifar10_functa}
\end{wraptable}

It is surprising that \functa underperform on both downstream tasks on CIFAR-10 given that \citet{functa} report competitive results on ShapeNet voxel classification and generation using relatively simple (MLP) architectures.

One key property of \functa latent representations is that they encode data as one global vector; i.e., each latent dimension encodes global features that span the entire image/data point when viewed in pixel space. We demonstrate this via a perturbation analysis on CIFAR-10, where changes in any latent dimension lead to global perturbations across the reconstructed image in pixel space, c.f. \cref{fig:cifar10_functa_perturbation}, \cref{sec:app:functa_limitations}. In contrast, in the data domain information is naturally spatially arranged and localized; indeed, architectures such as convolutional networks on pixel/voxel arrays or Transformers on spatio-temporal arrays explicitly exploit this spatial structure.

We hypothesize that the stark performance differences between \functa on CelebA / ShapeNet and CIFAR-10 arise due to differing complexities of the tasks at hand. For sufficiently simple tasks, architectures such as MLPs and Transformers are able to extract features that are useful for these tasks from global latent representations of the underlying neural field. For more challenging tasks however, they %
struggle to extract sufficiently expressive features.
This naturally motivates us to explore \emph{spatial} latent representations of functa, which we call \textit{\sfuncta}.

\section{Spatial Functa}
\label{sec:spatial_functa}

Here we present \textit{\sfuncta}, a generalization of functa that uses spatially arranged latent representations instead of flat latent vectors. This allows features at each spatial index to capture local information as opposed to global information spread across all spatial locations.
This simple change enables us to solve downstream tasks with more powerful architectures such as Transformers with positional encodings \citep{vaswani2017attention} and UNets \citep{Ronneberger2015}, whose inductive biases are suitable for spatially arranged data. In \cref{sec:experiments} we show that this greatly improves the usefulness of representations extracted from functa latents for downstream tasks, allowing the functa framework to scale to complex datasets such as ImageNet-1k at $256 \times 256$ resolution.

\textit{Functa} linearly map the (global) latent vector to a vector of shift modulations
, which is subsequently used to condition the neural field at all positions $\vx$, see \cref{fig:figure_1} \textit{(top left)}. Shift modulations are bias terms that get added to the activations at each layer the MLP neural field $f_\theta$. In \sfuncta we still use a vector of shift modulations to condition the neural field for each position $\vx$, however, the modulation vector $\vm_\vx$ is now position dependent. To achieve this, we first linearly map the spatial latent $\vz \in \mathbb{R}^{s\times  s\times c}$ to a spatially arranged shift modulation $\vm\in\mathbb{R}^{s \times s \times C}$ using a single convolutional layer $\vm = \lambda(\vz)$, where $s < d$, the image resolution. 
From the spatial shift modulation $\vm$ we then extract the position-dependent modulation vector $\vm_\vx\in\mathbb{R}^C$ by interpolation, see \cref{fig:figure_1} \textit{(top right)}. The pixel value output of the neural field $f_\theta$ queried at position $\vx$ is then given by $f_\theta(\vx; \vm_\vx)$.

The modulations $\vm_\vx$ are constrained to depend only on latent features that correspond to the coordinate of interest $\vx$, in order to enforce the property that $\vz$ encodes local information in space. We focus on two interpolation schemes:
(a) \textit{nearest neighbor} -- modulation vector is constant over patches in coordinate space; and (b) \textit{bilinear interpolation} of the $s\times s$ spatial features to obtain $\vm_\vx$.
Using nearest neighbor interpolation (with $\lambda$ a $1\times 1$ convolution) for grid data, such as images at $d \times d$ resolution, means that we have an independent $c$-dimensional latent feature per patch of size $d/s \times d/s$.
Also note that we recover the original \functa parameterization by setting $s=1$.

With the above parameterization of each data point as a spatial functa point, we train all parameters and build a (spatial) functaset exactly like \citet{functa} via meta-learning: we learn the parameters shared across all data points, namely the latent-to-modulation linear map $\lambda$ and the weights $\theta$ for the neural field $f_\theta$. After training, we freeze $\lambda$ and $\theta$ and create a functaset by converting each data point to a spatial latent $\vz$ via 3 gradient steps from a zero initialization.
We then use the spatial functaset $\{\vz^{(i)}\}_i$, with each data point represented by a latent $\vz^{(i)}$ as input data for downstream tasks.

\section{Experimental Results}
\label{sec:experiments}

In the following we train and build spatial functasets for CIFAR-10 and ImageNet256 with differing latent shapes and interpolations. To assess their quality, we first report reconstruction PSNRs on the respective test sets. We then train classifiers and generative models on these functasets, evaluating performance via: (a) top-1 classification test accuracies using Transformer classifiers; and (b) FID scores using convolutional UNet-based diffusion models.

\subsection{CIFAR-10 classification and generation with Spatial Functa}
\label{sec:exp_cifar10}
\begin{table}[h]
\centering
\setlength{\arrayrulewidth}{0.3mm}
\begin{tabular}{c c C C C C}
\toprule
Latent shape & Interpolation & \text{Test PSNR } \uparrow & \text{Top-1 acc } \uparrow & \text{FID (uncond) } \downarrow & \text{FID (cond) } \downarrow \\\midrule
$8\times8\times16$ & 1-NN & 37.2 \dB & \mathbf{90.3\%} & \mathbf{16.5} & 11.2* \\ 
$8\times8\times16$ & Linear & 35.6 \dB & 83.0\% & 25.6 & 11.6* \\ 
$4\times4\times64$ & 1-NN & \mathbf{39.0 \dB} & 88.2\% & 44.6 & \mathbf{10.4*} \\
$4\times4\times64$ & Linear & 35.3 \dB & 80.8\% & 71.2 & 11.0* \\
\bottomrule
\end{tabular}
\caption{CIFAR-10 spatial functa results. *Model memorizes subset of training set. Details in \ref{sec:app:cifar10_memorization}.}
\label{tab:cifar10_spatial_functa}
\end{table}

We first revisit CIFAR-10 to show that spatial functa overcome the limitations discussed in \Cref{sec:functa_limitations}. 
Instead of a single feature vector, we use a grid of smaller feature vectors such that the overall dimensionality of the latent stays the same (1024 dims). See \cref{sec:app:cifar10_exp_details} for model and training details.

\begin{figure}[h]
    \centering
    \begin{subfigure}[t]{0.5\textwidth}
        \centering
        \adjincludegraphics[trim={0 0 0 {0.5\height}},clip, width=\textwidth]{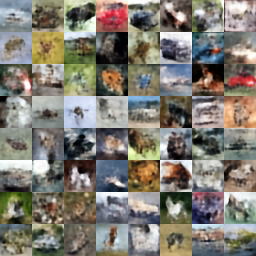}
    \end{subfigure}%
    ~ 
    \begin{subfigure}[t]{0.5\textwidth}
        \centering
        \adjincludegraphics[trim={0 0 0 {0.5\height}},clip, width=\textwidth]{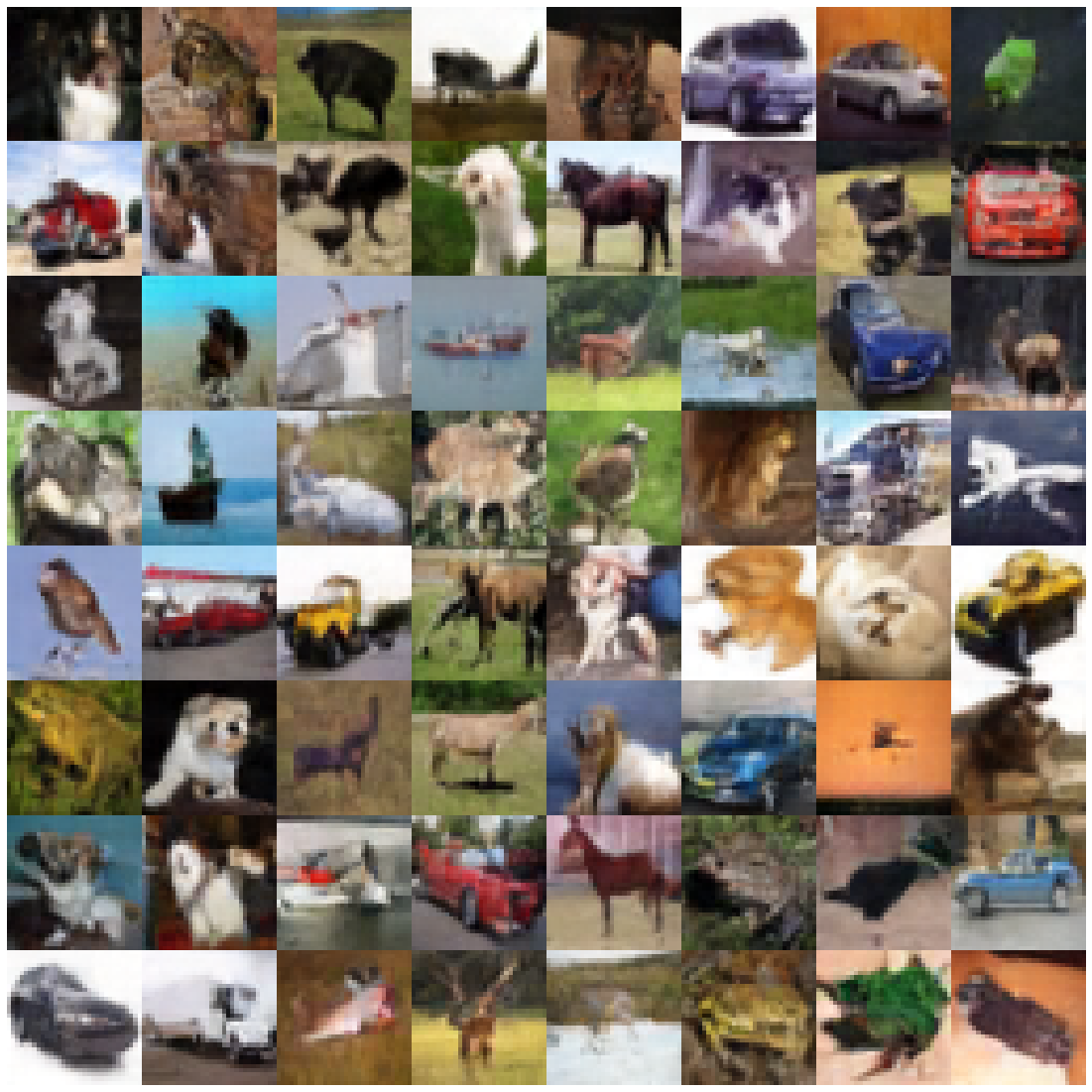}
    \end{subfigure}
    \caption{Uncurated samples from uncondtional DDPM with best FID score. \textit{left:} original vector-valued functa, $\FID=78.2$; \textit{right:} spatial functa, $\FID=16.5$.}
    \label{fig:cifar10_functa_samples}
\end{figure}

On both tasks, \sfuncta perform markedly better than the original vector-valued functa despite their similar reconstruction PSNR: For classification we now achieve a top-1 accuracy of $90.3\%$ (see \cref{tab:cifar10_spatial_functa}) instead of $68.3\%$ (see \cref{tab:cifar10_functa}). Similar improvements can be observed for image generation with Latent Diffusion: we achieve FIDs of $16.5$ (with an unconditional model without classifier-free guidance) and $10.4$ (with a conditional model with classifier-free guidance) for \sfuncta (see \cref{tab:cifar10_spatial_functa}) compared to $134.8$ for functa with the same latent dimensionality and $78.2$ with fewer latent dimensions (see \cref{tab:cifar10_functa}). In \cref{fig:cifar10_functa_samples} we contrast uncurated samples of vector-valued and spatial functa.

In general, nearest neighbor interpolation tends to perform better than linear interpolation for downstream tasks. In both reconstructions and samples we do not observe patch artifacts, likely because the reconstruction PSNR is high enough, so the diffusion model learns to generate samples that match up on the patch boundaries. Moreover, the higher the spatial latent dimensionality (while keeping total number of latent dimensions fixed), the better the performance on both downstream tasks. This observation indicates the importance of locality for extracting useful features in downstream tasks through Transformers and UNets. 

Similar to \citet{nichol2021improved} we observed that class-conditional diffusion models can easily collapse to memorizing (a subset of) the CIFAR-10 training set. 
We discuss this overfitting and our solution -- using unconditional DDPM on the dataset instead -- in \cref{sec:app:cifar10_memorization}.

\subsection{ImageNet $256\times256$ classification and generation with Spatial Functa}
\label{sec:exp_imagenet}

We repeat similar experiments on ImageNet-1k $256\times256$. 
We again found that nearest neighbor interpolation always gave better performance and that larger spatial dimensions of the latents yielded the best PSNR, which also correlates with best top-1 accuracy and best FID scores, see \cref{tab:imagenet_spatial_functa}. 
See \ref{sec:app:imagenet_exp_details} for further results and details on architectures and hyperparameters for these downstream tasks.

\begin{table}[h]
\centering
\setlength{\arrayrulewidth}{0.3mm}
\small{
\begin{tabular}{c c c c c}
\toprule
Model & Input shape& Test PSNR $\uparrow$ & Top-1 acc $\uparrow$ & FID (cond) $\downarrow$ \\
\midrule
\multirow{9}{*}{Spatial Functa (1-NN)} & $8\times8\times256$ & 28.3dB & 76.6\% & 17.9 \\ 
& $8\times8\times512$ & 30.6dB & 76.5\% & 23.5 \\
& $8\times8\times1024$ & 25.7dB & 76.5\% & - \\
& $16\times16\times64$ & 28.9dB & 80.4\% & 12.5 \\ 
& $16\times16\times128$ & 37.8dB & \textbf{80.7\%} & 10.5 \\
& $16\times16\times256$ & 37.2dB & 80.6\% & 11.7 \\
& $32\times32\times16$ & 28.6dB & - & 12.4 \\ 
& $32\times32\times32$ & 31.7dB & - & 10.5 \\ 
& $32\times32\times64$ & 37.7dB & - & 8.8 \\
& $32\times32\times64$* & \textbf{38.4 dB} & - & \textbf{8.5} \\
\midrule
\multirow{2}{*}{ViT-B/16} & $224\times224\times3$ & - & 79.8\% & - \\
 & $384\times384\times3$ & - & \textbf{81.6\%} & - \\
\midrule
LDM-8-G & $32\times32\times4$ (14 bits) & 23.1dB & - & 7.8 \\
LDM-4-G & $64\times64\times3$ (13 bits) & 27.4dB & - & \textbf{3.6} \\
\bottomrule
\end{tabular}
}
\caption{ImageNet classification and diffusion results. *Indicates that a $3\times3$ Conv was used instead of $1\times1$ Conv for the latent to modulation linear map.}
\label{tab:imagenet_spatial_functa}
\end{table}

For classification, there is little difference across different numbers of feature dimensions.
Note that we omit classification results for $32\times 32$ latent dimensions as this would correspond to a sequence length of $1024$ for the Transformer classifier, which makes the vanilla Transformer too costly in terms of memory. For a similar reason ViT on ImageNet only explores patch sizes $\geq 32$ \citep{dosovitskiy2020image, steiner2021train}. We also omit diffusion results for latent shape $8\times8\times1024$ as DDPM suffered from training instabilities.

Our best classification performance using \sfuncta is 80.7\% on $16\times16\times128$ latent dimensions, which is competitive with ViT-B/16, the best performing Vision Transformer at patch-size 16 (i.e., using the same sequence length of 256 as our best performing Transformer) trained on the same ImageNet-1k dataset with similar data augmentation \footnote{ViT actually uses stronger augmentations, RandAugment and MixUp, whereas we only use RandAugment} \citep{steiner2021train}. Note that the numbers for ViT are for two different resolutions $224\times 224$ and $384\times 384$; our resolution of $256 \times 256$ is closer to the former yet has better accuracy. Moreover our classifier has 64M parameters compared to 86M parameters for ViT-B/16.

For image generation, the best model achieves an FID of 8.50 (with 277M parameters); see curated samples in \cref{fig:figure_1} \textit{(bottom right)}, and uncurated samples in \cref{fig:imagenet_uncurated_samples}. This FID score is close to the best performing Latent Diffusion model with the same spatial downsampling factor: LDM-8-G at 7.76 (with 506M parameters) \citep{rombach2022high}. We expect better FID with a smaller downsampling factor, just as LDM-4-G (3.60) is better than LDM-8-G (7.76).

\subsection{Ablations}
\label{sec:ablations}

Besides model size and latent shape, several other hyperparameters affect model performance, in particular for image generation. These include the normalization of the functaset, the guidance scale for classifier-free guidance, and the sampling of timesteps for diffusion training. We discuss these ablations as well as preliminary results for compression of the spatial functa via quantization in \ref{sec:app:ablations}.

\section{Related Work}
Here we focus on works that are relevant to spatial functa or date after \citet{functa}. Please refer to their related work section for references on functa that precede it.

Modeling neural fields with spatial latent features has become popular across a wide range of modalities. Grounding outputs of neural fields on spatial latents (with or without interpolation) has been shown to allow for efficient fitting and inference of neural fields at high fidelity in modalities such as 3D scenes/shapes \citep{jiang2020local,chabra2020deep,chibane2020implicit,yu2021plenoxels,mueller2022instant,chen2022tensorf},
videos \citep{kim2022scalable} and images \citep{mehta2021modulated,chen2021learning}. We show that such spatial representations of neural fields are also particularly important for deep learning on neural fields to solve both discriminative (e.g.\ classification) and generative (e.g.\ synthesis) downstream tasks at scale. Although we demonstrate this only on image datasets (up to ImageNet-1k at $256\times256$ scale), we expect this observation to also hold more generally for sufficiently complex datasets of other modalities.

Two recent works further exemplify the importance of spatial latents for generative modeling. Using diffusion they successfully model 3D shapes \citep{shue20223d} and scenes \citep{bautista2022gaudi} and crucially rely on spatial representations and a functa-like framework that consists of two decoupled stages: 1. creation of a dataset of neural fields and 2. subsequent downstream learning on this dataset.
Both works focus on generative downstream tasks with diffusion models and on datasets with a small number of data points ($\leq 10$k shapes/scenes). In contrast, our work evaluates on both generative \emph{and} discriminative tasks and uses datasets with ${}>1$M data points (ImageNet).

Concurrently, \citet{dl_inr_shapes2023} employ the functa framework for 3D shapes, namely treating each shape as a neural field, and do not rely on spatial latent features, rather a single latent vector representation per shape. The key difference is that they use an encoder to map the weights and biases of each SIREN field to the latent representation, whereas in functa the encoding is done by gradient updates. These latents are shown to be useful for various downstream tasks such as point cloud retrieval, classification, generation and segmentation. However the approach has only been applied to ShapeNet, where the original functa methodology was also shown to perform well on downstream tasks, so it remains to be seen whether their proposed methodology can be extended to CIFAR-10 and ImageNet scale. 

Several recent works propose continuous kernel convolutions \citep{romero2021ckconv,romero2021flexconv} that can process functional representations, and \citet{wang2022deep} employ such architectures for deep learning on neural fields. However as it stands, this approach does not scale to datasets beyond CIFAR-10 due to the compute/memory costs of continuous kernel convolutions.

\section{Conclusion and Future Work}
In this work, we have shown that the original neural field parameterization used in \emph{\functa} \citep{functa} struggles to scale functa classification and generation to even relatively small datasets such as CIFAR-10. We then proposed a spatial parameterization, termed \emph{\sfuncta}, and showed that this allows for scaling functa to ImageNet-1k at $256 \times 256$ resolution, achieving competitive classification performance with ViT \citep{steiner2021train} and generation performance with Latent Diffusion \citep{rombach2022high}.

We emphasize that achieving competitive results on image tasks is not the ultimate goal, but rather a stepping stone for scaling up functa in other modalities such as video (+ audio) and static/dynamic scenes. We believe these higher dimensional modalities are where the functa framework will truly shine with scale, since the vast amounts of redundant information in array-representations of these modalities can be captured much more efficiently as neural fields.

We also highlight that while we can successfully scale functa with spatial functa, we have not fully answered the question as to why the original vector-valued functa representations are difficult to use in larger scale downstream tasks. This is an interesting open problem, and a potential avenue for exploration would be to devise efficient architectures that allow extracting useful representations from flat vector representations of functa for sufficiently complex datasets at scale.

\subsubsection*{Acknowledgments}
We thank Yee Whye Teh for helpful discussions around the project, and S. M. Ali Eslami for helpful feedback regarding the manuscript.

\bibliography{main}
\bibliographystyle{iclr2023_conference}

\clearpage
\appendix
\section{Background on Functa}
\label{sec:app:background}
In this section we give a summary of the methodology used by \cite{functa}. Please refer to the original paper for more details.

\textit{Functa} are used to describe neural fields $f_\theta$ that should be thought of as data. In this work, each $f_\theta$ is a SIREN \citep{sitzmann2020implicit}, namely an MLP with sine non-linearities, that is fit to a single image by minimizing reconstruction loss on all pixels. In order to perform deep learning tasks on this dataset of neural fields (\textit{functaset}), we would like to represent each field (\textit{functum}) with an array that can be used as an input to neural networks.

A na{\"i}ve approach would be to use the parameter vector $\theta$ of the MLP as the array-representation of the field, but this representation is likely unnecessarily high dimensional.
A more economical approach would be to share parameters across different fields in order to amortize the cost of learning the shared structure across these fields, and model the variations with a lower dimensional, per-field representation that modulates the shared parameters. One can split the parameters as follows: share the weights of a base SIREN across all fields and then infer per-field biases or \textit{shifts}. The concatenation of these bias terms across all SIREN layers is referred to as the \textit{shift modulation} (see \cref{fig:figure_1}), which now serves as a more compact representation of each field. \cite{functa} show that only adapting the biases is already sufficient to model the variations across different fields to high reconstruction accuracy.

One can further reduce the dimensionality of these field representations by having a latent vector $\vz$ that is linearly mapped onto the shift modulation; the linear map is shared across all fields. These latents then serve as representations for various downstream deep learning tasks such as classification or generation used in \cite{functa}.

In order to learn the shared parameters (base SIREN and latent-to-modulation linear map) and the per-field representation $\vz$, \citet{functa} propose to use meta-learning, namely a double-loop optimization procedure that minimizes the reconstruction loss on all pixels. Given a batch of $B$ images, the inner loop initializes $B$ copies of per-field latents $\vz$ to zero, and updates each with three SGD steps with respect to the reconstruction loss. The outer loop updates the shared parameters with Adam \citep{kingma2014adam} to minimize the reconstruction loss using the inner-loop updated values of $\vz$, with gradients flowing through the inner loop. \Citet{functa} show that this allows one to effectively learn the shared parameters, after which the $\vz$ for each field can be fit very fast using \textit{three} SGD inner loop steps. They show that three steps is already sufficient to achieve high quality reconstructions. This is how a \textit{functaset} is created from each dataset.

\section{Limitations of naively scaling functa}
\label{sec:app:functa_limitations}

\subsection{Experimental details for functa on CIFAR-10}
\label{sec:app:functa_cifar10}

\paragraph{Meta-learning} To create the CIFAR-10 functaset (original vector functa), we use the same hyperparameter configurations as the meta-learning used in \cite{functa} on CelebA-HQ $64\times64$ except for the total batch size. Namely, we use $\omega_0=30$, SIREN width $512$, depth $15$, outer learning rate 3e-6, using meta-SGD, with total batch size $128$. We use the same initializations for the SIREN weights and biases, meta-SGD learning rates, and a zero initialization of the latent $\vz$.

\paragraph{Classification} We use the same classifier architecture as \cite{functa}, namely an MLP with SiLU/swish activations and dropout, and also test a vanilla Transformer (with positional encodings). We additionally use:
\begin{itemize}
    \item Standard data augmentation with left/right flips and random crops (taking $32\times32$ crops of a $40\times40$ image obtained by padding zeros to the boundary of the original image), using 50 randomly sampled augmentations per training image to create the augmented functaset.
    \item Scalar normalizing factor to normalize the functa latents before feeding them into the classifier.
    \item Label smoothing \citep{szegedy2016rethinking} hyperparameter $l$ such that e.g. the one hot label $[1,0,\ldots,0]$ for $n$ classes is replaced with $[1-l+ l/n, l/n, \ldots, l/n]$.
    \item Weight decay with AdamW \citep{loshchilov2017fixing}.
    \item Track exponential moving average of parameters throughout training that is used at evaluation time, using decay rate $0.9999$.
\end{itemize}
We performed a large hyperparameter sweep over normalizing factor in $[0.01, 0.03, 0.1]$, learning rate in $[1\text{e-}4, 1\text{e-}3, 1\text{e-}2]$, label smoothing $l$ in $[0, 0.1]$, weight decay in $[0, 1\text{e-}1]$, dropout in $[0, 0.1, 0.2]$, MLP width in $[128, 256, 512, 1024]$, MLP depth in $[2,3,4,5]$, Transformer width in $[64, 128, 256]$, Transformer depth in $[2,3,4]$, with Transformer ffw-width being double the width. All models were trained for 300k iterations on a batch size of $256$, and we report the best test accuracy throughout training. The optimal performance was given by an MLP on $512$-dimensional latents at $68.3\%$ test accuracy with normalizing factor $0.01$, learning rate $1\text{e-}3$, label smoothing $0.1$, weight decay $0.1$, dropout $0.2$, width $1024$, depth $3$. The best performing Transformer was on $256$-dimensional latents at $63.5\%$ test accuracy with normalizing factor $0.1$, learning rate $1\text{e-}3$, label smoothing $0.1$, weight decay $0.1$, dropout $0.1$, width $256$, depth $2$.

\paragraph{Generation with diffusion models} We use the same DDPM \citep{Ho2020} architecture as \cite{functa} with ResidualMLP blocks, following their hyperparameters configurations and sweeps as closely as possible: functa latents are mean centered and normalized by the elementwise mean and standard deviation across the training set, and Adam is used with a learning rate schedule that warms up linearly from $0$ to $3\text{e-}4$ for the first $4000$ iterations, followed by inverse square root decay. We use a batch size of $128$ and sweep over the width in $[512, 1024, 2048, 4096]$, number of blocks in $[2,4]$, training for 1M iterations. The optimal performance was given by $256$-dimensional latents at $78.2$ FID with width $4096$, number of blocks $2$.

\subsection{Functa reconstructions on CIFAR-10}

\begin{figure}[h]
	\centering
	\begin{subfigure}[t]{0.5\textwidth}
		\centering
		\includegraphics[width=0.9\textwidth]{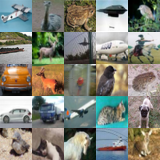}
		\caption{Original test images}
	\end{subfigure}%
	~ 
	\begin{subfigure}[t]{0.5\textwidth}
		\centering
		\includegraphics[width=0.9\textwidth]{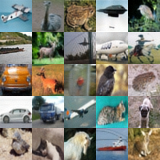}
		\caption{Reconstructions from functa}
	\end{subfigure}
	\caption{\textit{Functa} with flat (vector) latent representations obtained through meta-learning can reconstruct test images to high fidelity. Compare the riginal CIFAR-10 test images \textit{(left)} and their functa reconstructions \textit{(right)}.}
	\label{fig:cifar10_functa}
\end{figure}

In \cref{sec:functa_limitations} we showed that naively scaling vector-valued \functa \citep{functa} does not lead to representations that can be easily used for downstream tasks despite yielding high reconstruction PSNRs. In \cref{fig:cifar10_functa} we show original images from the CIFAR-10 test set as well as their high-quality reconstructions using vector-valued \functa.
 
\subsection{Functa Perturbation Analysis on CIFAR-10}

\begin{figure}[h]
	\centering
	\begin{subfigure}[t]{0.5\textwidth}
		\centering
		\includegraphics[width=\textwidth]{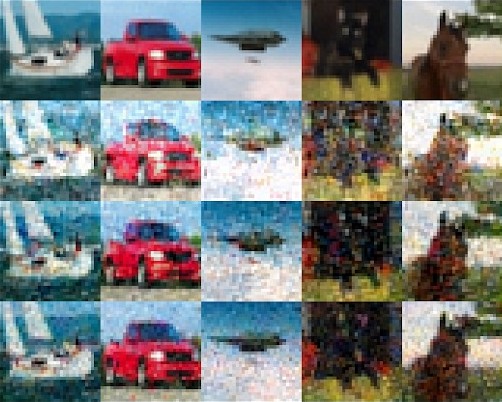}
		\caption{Functa reconstructions}
	\end{subfigure}%
	~ 
	\begin{subfigure}[t]{0.5\textwidth}
		\centering
		\includegraphics[width=\textwidth]{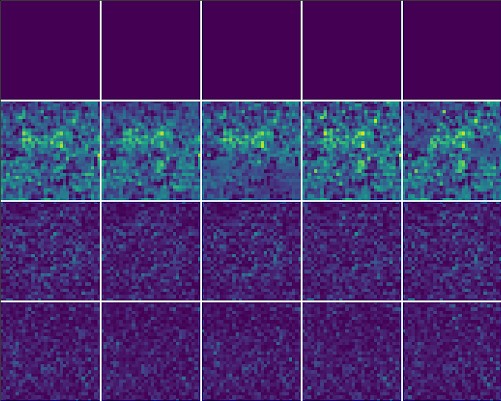}
		\caption{Mean absolute error in pixel space}
	\end{subfigure}
	\caption{Perurbation of vector-valued \functa on CIFAR-10. \textit{Row 1:} unperturbed functa reconstructions. \textit{Row 2:} perturbing latent dimension with largest mean absolute error in pixel space (MAE). \textit{Row 3:} perturbing latent dimension with median MAE. \textit{Row 4:} perturbing latent dimension with lowest MAE.}
	\label{fig:cifar10_functa_perturbation}
\end{figure}

To show that the individual dimensions of the latent vector indeed affect the entire image (as opposed to local regions) we perform a simple perturbation analysis similar to \citet{functa}, see \cref{fig:cifar10_functa_perturbation}. For a set of functa we perturb the same latent dimension and then observe the changes in the pixel space. As discussed in the main text, perturbing each latent dimension individually gives indeed rise to perturbations throughout the entire image.

\section{Spatial Functa: experimental details and additional results}
\label{sec:app:exp}

\paragraph{Coordinate representation for SIREN} When using the \emph{nearest neighbor} interpolation for spatial functa, where the $c$-dimensional latent feature vector at each spatial index represents an image patch, we found that using a per-patch coordinate representation as the input to the SIREN neural field helped optimization such that reconstructions achieve better PSNR values compared to a global coordinate representation. For example, for the bottom-left patch of size $d/s \times d/s$, we use coordinate representations $\vx \in [0, 1]^2$ rather than $\vx \in [0, s/d]^2$. This allowed the meta-learning step to fit individual images to higher PSNR. For \emph{linear} interpolation we empirically found that using a binary representation of the global coordinates gave better PSNR values than the patch-based coordinate representation above. For example, for CIFAR-10 at $32 \times 32$ resolution, the coordinate at location $(3/32, 5/32)$ would be converted to binary $(00011, 00101)$, concatenated to $0001100101$, then converted to a binary vector $[0,0,0,1,1,0,0,1,0,1]$. Hence for an image of $2^m \times 2^m$ resolution, we'd use a $2m$-dimensional binary coordinate representation. Although we did not try using the binary representation for \emph{nearest neighbor} interpolation, we expect it to perform similarly to the per-patch coordinate representation.

\paragraph{Classification} We use the standard vanilla pre-LN Transformer \citep{vaswani2017attention, wang2019learning, xiong2020layer} with absolute positional encodings where each input token is the $c$-dimensional latent feature vector of the spatial functa at each spatial index. Hence the sequence length of the Transformer is $s^2$ where each spatial functa is given by latent $\vz \in \mathbb{R}^{s\times s\times c}$. Similar to classification experiments on the original vector-valued functa in \cref{sec:app:functa_cifar10}, we use data augmentation, scalar normalization of input latents, label smoothing and weight decay. Dropout is only applied once just before the final linear layer that maps onto the logits. For CIFAR-10, we use the same random crop and flip augmentations, whereas for ImageNet we use random distorted bounding box crops and flips (following \cite{szegedy2016rethinking}) followed by RandAugment \citep{cubuk2020randaugment} with $2$ sequential augmentations at magnitude $5$.

\paragraph{Generation with diffusion models} We closely follow the improved implementation of discrete-time denoising diffusion probabilistic models (DDPMs) \citep{Ho2020} by \citet{dhariwal2021diffusion}, and optionally use classifier-free guidance \citep{ho2022classifier} for generation. We use $1000$ diffusion time steps and a cosine noising schedule and train the denoising model using the original (simplified) DDPM loss though we also explored non-uniform sampling distributions for the time steps. Our denoising architecture is the same UNet \citep{Ronneberger2015} used by \citet{dhariwal2021diffusion} and we use the $\epsilon$ (noise prediction) parameterization, which we found to work best.

\subsection{Experimental details for CIFAR-10}
\label{sec:app:cifar10_exp_details}

\paragraph{Meta-learning} For meta-learning, we sweep over the below hyperparameters to find the ones that give optimal PSNR on the test set, using $\omega_0=10$, batch size $128$, 200k training iterations for $8 \times 8$ spatial latent dimensions and 500k iterations for $4 \times 4$ spatial latent dimensions:

\begin{table}[h]
    \centering
    \begin{tabular}{Cc | CCc}
    \toprule
     \textbf{latent shape} & \textbf{interpolation} & \textbf{SIREN width} & \textbf{SIREN depth}& \textbf{outer learning rate} \\\midrule
     8\times8\times16 & $1$-NN & 256 & 6 & 3e-5\\
     8\times8\times16 & linear & 256 & 6 & 1e-4\\
     4 \times 4 \times 64 & $1$-NN & 256 & 10 & 3e-5\\
     4 \times 4 \times 64 & linear & 256 & 10 & 3e-5\\
     \bottomrule
\end{tabular}
    \caption{Optimal hyperparameter configuration for meta-learning on CIFAR-10}
    \label{tab:app:spatial_cifar10_meta_learning_hypers}
\end{table}

\paragraph{Classification} We sweep over the below hyperparameters to find the ones that give optimal test accuracy, using label smoothing $0.1$, weight decay $0.1$, $16$ Transformer heads, for 100k training iterations, see \cref{tab:app:spatial_cifar10_classification_hypers}.

\renewcommand*\rottab{\multicolumn{1}{R{30}{1.5em}}}
\begin{table}[h]
    \centering
    \begin{tabular}{Cc | CCCCcCC}
    \toprule
     \textbf{latent shape} & \textbf{interpolation} & \rottab{normalization scale} & \rottab{Transformer width}&\rottab{ffw width} & \rottab{number of blocks} & \rottab{learning rate} & \rottab{dropout rate} & \rottab{batch size} \\\midrule
     8\times8\times16 & $1$-NN      & 0.08  & 128   & 256   & 12    & 1e-2  & 0.1   & 256\\
     8\times8\times16 & linear      & 0.1   & 128   & 256   & 12    & 1e-2  & 0.2   & 256\\
     4 \times 4 \times 64 & $1$-NN  & 0.01  & 384   & 768   & 12    & 2e-3  & 0.1   & 512\\
     4 \times 4 \times 64 & linear  & 0.01  & 384   & 768   & 16    & 2e-3  & 0.2   & 512 \\
     \bottomrule
\end{tabular}
    \caption{Optimal hyperparameter configuration for classification on CIFAR-10}
    \label{tab:app:spatial_cifar10_classification_hypers}
\end{table}

\paragraph{Generation} For generation with diffusion models we swept over several architecture and noising hyperparameters to determine the best configuration for each latent space shape and interpolation based on FID. 
We trained for $500$k iterations using a batch size of $256$ and learning rate $10^{-4}$ using Adam with default parameters, dummy-label proportion of $0.2$ for classifier-free guidance. We used vector-valued normalization and tune the scale factor (see \cref{sec:app:ablations:normalization}). See \cref{tab:app:spatial_cifar10_hypers} for the optimal hyperparameters in terms of the denoising architecture.

\renewcommand*\rottab{\multicolumn{1}{R{30}{1.5em}}}

\begin{table}[h]
    \centering
    \begin{tabular}{cCc | CcCCCCC}
    \toprule
     \textbf{conditional} &\textbf{latent shape} & \textbf{interpolation} & \rottab{normalization scale} & \rottab{noising schedule}&\rottab{\# res blocks} & \rottab{channels} & \rottab{channel multiplier} & \rottab{attention res} & \rottab{dropout rate} \\\midrule
     \xmark & 8\times8\times16 & $1$-NN & 2 & linear  & 1 & 320 & (1, 1) &  (4,) & 0.4\\
     \xmark & 8\times8\times16 & linear & 2& cosine  & 2 & 320 & (2, 2) & (2, ) & 0.4\\
     \xmark & 4 \times 4 \times 64 & $1$-NN & 2 & cosine & 2 & 320 & (2, 2) & - & 0.4 \\
     \xmark & 4 \times 4 \times 64 & linear & 2 & cosine & 2 & 320 & (2, 2) & (4,) & 0.4 \\
     \cmark & 8\times8\times16 & $1$-NN & 2 & cosine & 3 & 320 & (2, 2, 2) &  (8,) & 0.1\\
     \cmark & 8\times8\times16 & linear & 2 & cosine & 3 & 320 & (2, 2, 2) & (8, 4, ) & 0\\
     \cmark & 4 \times 4 \times 64 & $1$-NN & 2 & cosine & 3 & 320 & (2, 2) & (4,) & 0 \\
     \cmark & 4 \times 4 \times 64 & linear & 2 & cosine & 3 & 320 & (2, 2, 2) & (4, 2) & 0 \\
     \bottomrule
\end{tabular}
    \caption{Optimal hyperparameter configuration for diffusion models on CIFAR-10. Note that all conditional models overfit and memorize a subset of the training set, see \protect\cref{sec:app:cifar10_memorization}.}
    \label{tab:app:spatial_cifar10_hypers}
\end{table}

\subsection{Assessing memorization on CIFAR-10}
\label{sec:app:cifar10_memorization}

\begin{figure}[h]
	\centering
	\begin{subfigure}[t]{0.35\textwidth}
		\centering
		\begin{tikzpicture}
        \node[inner sep=0em, outer sep = 0em] (x) {\adjincludegraphics[trim={0 0 0 {0.5\height}},clip, width=0.8\textwidth]{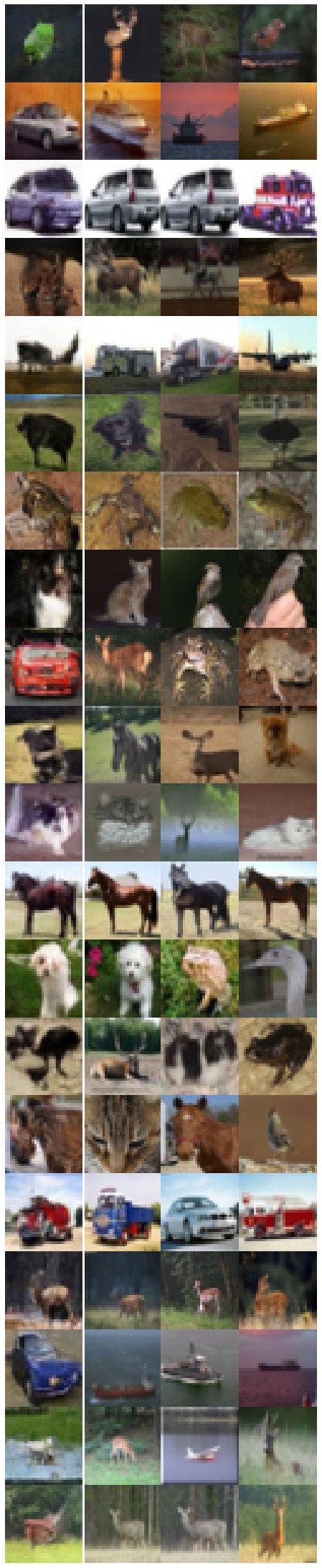}};
        \node (left) at (x.north west){};
        \node (right) at (x.north east){};
        \node[anchor=south] at ($(left)!1/8!(right)$) {Sample};
        \node[anchor=south] at ($(left)!5/8!(right)$) {Nearest neighbors};
        \end{tikzpicture}  
	    \caption{unconditional diffusion model}
	\end{subfigure}%
	~
	\begin{subfigure}[t]{0.35\textwidth}
		\centering
    	\begin{tikzpicture}
        \node[inner sep=0em, outer sep = 0em] (x) {\adjincludegraphics[trim={0 0 0 {0.5\height}},clip, width=0.8\textwidth]{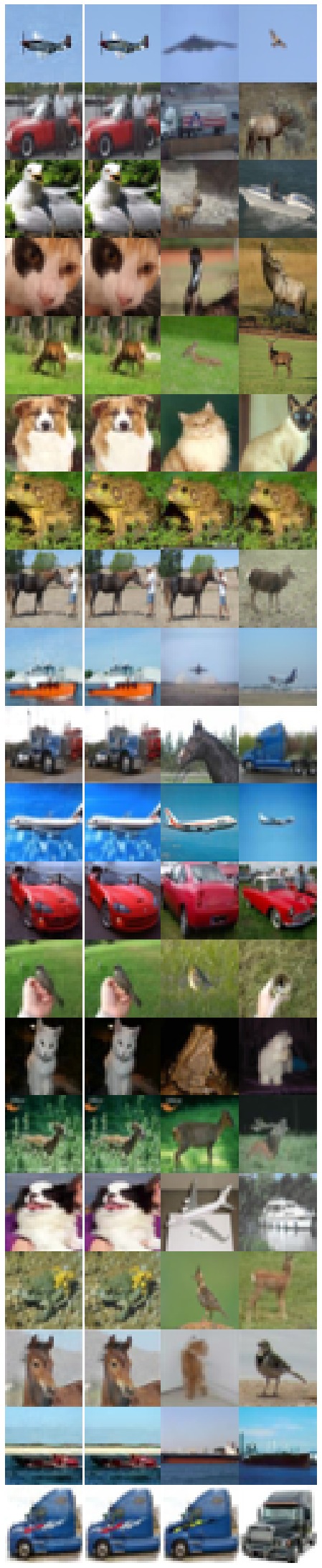}};
        \node (left) at (x.north west){};
        \node (right) at (x.north east){};
        \node[anchor=south] at ($(left)!1/8!(right)$) {Sample};
        \node[anchor=south] at ($(left)!5/8!(right)$) {Nearest neighbors};
        \end{tikzpicture}  
	    \caption{class-conditional diffusion model}
	\end{subfigure}
    \caption{Assessing memorization of CIFAR-10 training set. For each sample we show the three nearest neighbors (in pixel-space) from the training set.}
    \label{fig:cifar10_memorization}
\end{figure}
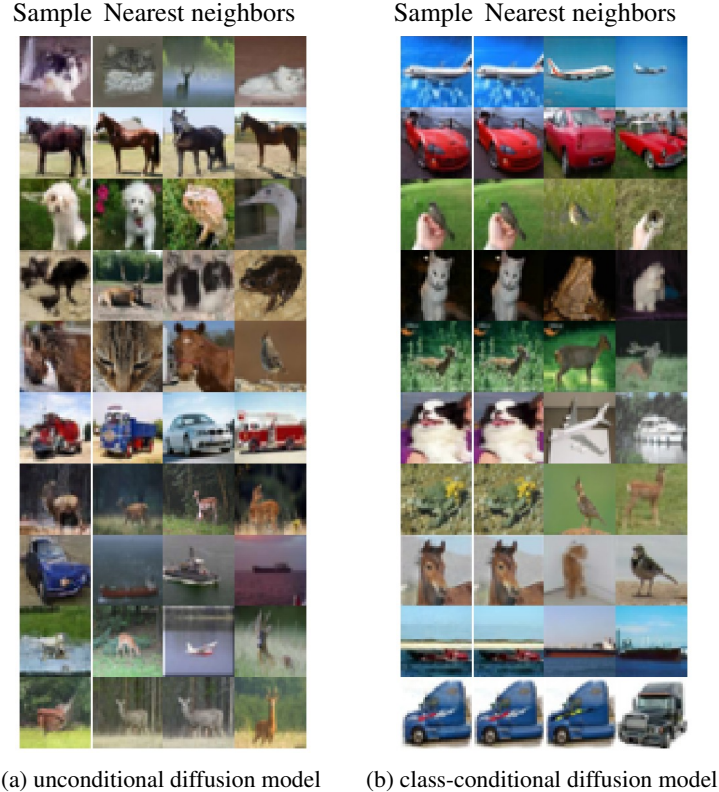

In \cref{sec:exp_cifar10} we mentioned that class-conditional diffusion models can overfit on CIFAR-10 by memorizing a subset of the training data; i.e. at sampling time the model only produces samples from this subset. Here, we discuss this in more detail.

We assessed memorization by finding the closest training set image (in pixel-space) to each model sample, see \cref{fig:cifar10_memorization} for examples. We found that diffusion models with class-conditional UNets of different sizes perfectly memorized a subset of the training data, see \cref{fig:cifar10_memorization}(b). While \citet{nichol2021improved} found that using a suitable dropout rate in the UNet allowed for a good trade-off between sample quality and overfitting for diffusion on pixel space, we were unable to find such a trade-off for our models -- they overfit even with strong dropout or yielded low-quality samples. Instead, we found that using unconditional models provided a better trade-off, see \cref{fig:cifar10_memorization}(a).

Despite memorizing training examples almost perfectly with class-conditional models, we still observe a relatively large FID of $\sim 10$. Below we show that this is due to mode dropping; i.e., the model does not memorize the entire training set but only a subset, thus not matching the covariance of the training data used for the FID evaluation.

We sampled $5000$ images per class from the model and looked up their nearest neighbors in the training set. We also verified that each sample indeed corresponded to a very close example in the training set. We then counted the number of \textit{distinct} samples per class such found. If the model perfectly remembered the training data, it should produce each image of a particular class from the training set with the same probability ($1/5000$ in the case of the CIFAR-10). The expected number of distinct samples when sampling $5000$ images from the class would then be $3161$ (with a standard deviation of $34$); this corresponds to drawing $5000$ times i.i.d. from a set of $5000$ distinguishable objects \textit{with} replacement and asking how many unique objects we would draw. We find that the overfitting models produce significantly fewer unique samples, see~\cref{tab:app:cifar10_memorization}. This corresponds to mode dropping, i.e. memorization of only a subset of the training images, or at least sampling them with very uneven probabilities.

\begin{table}[ht]
	\centering
	\begin{tabular}{C | CCCCCCCCCC}
		\toprule
		\text{\textbf{expected}} & \multicolumn{10}{c}{\textbf{observed} (class number)}\\
		& (1) & (2) & (3) & (4) & (5)& (6) & (7) & (8) & (9) & (10) \\\midrule
		3161 & 1219 & 1927 & 1800 & 1904 & 1547 & 1437 & 1492 & 1644 & 1555 & 1842 \\
		\bottomrule
	\end{tabular}
	\caption{Counts of unique images memorized from the training set for each class vs the expected number if samples were drawn i.i.d. with replacement when sampling $5000$ times.}
	\label{tab:app:cifar10_memorization}
\end{table}

\subsection{Experimental details for ImageNet $256 \times 256$}
\label{sec:app:imagenet_exp_details}
For ImageNet-1k $256 \times 256$, we show the hyperparameter configurations for the spatial functa that give the best classification performance and those that give the best generation performance.

\textbf{Meta-learning} We use the following hyperparameter configurations for each of these latent shapes and choice of interpolation:

latent shape $16\times16\times128$, 1-NN interpolation: batch size $16$, 500k training iterations, SIREN width $256$, depth $12$, $\omega_0=20$, outer learning rate 1e-5.

latent shape $32\times32\times64$, 1-NN interpolation, using $3\times3$ Conv for latent to modulation mapping: batch size $16$, 500k training iterations, SIREN width $256$, depth $8$, $\omega_0=10$, outer learning rate 3e-5.

\textbf{Classification} The following hyperaparameters gave the best classification results for latent shape $16\times16\times128$: label smoothing $0.1$ , weight decay $0.2$, $200$ training epochs, scalar normalization $0.005$, learning rate 1e-3, dropout $0.2$, batch size $1024$, Transformer width $768$, ffw width $1536$, number of blocks $18$, $16$ heads.

\textbf{Diffusion} The following hyperaparameters gave the best FID score for latent shape $32\times32\times64$: 1.5M training iterations, batch size $256$, learning rate 1e-4, dummy-label proportion $0.2$, vector-valued normalization with scale factor $2.5$, no dropout, cosine noising schedule, $2$ residual blocks, channels $320$, channel multiplier $(1, 2, 2, 2)$, attention resolutions $(8, 16)$.

\begin{figure}[h]
    \centering
    \includegraphics[width=\textwidth]{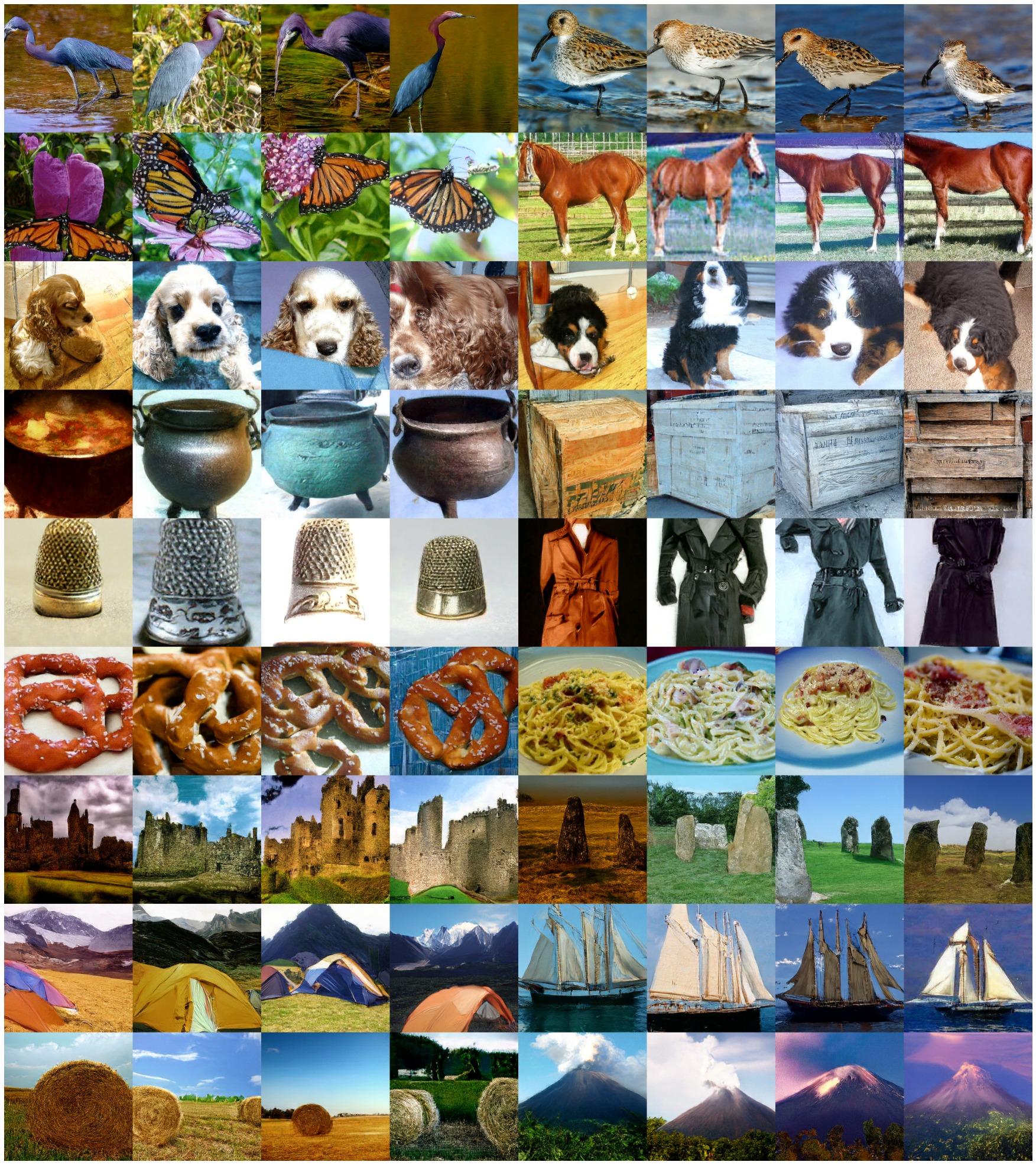}
    \caption{Uncurated ImageNet $256\times256$ samples at guidance scale=5 for different classes from the best diffusion model trained on spatial functa.}
    \label{fig:imagenet_uncurated_samples}
\end{figure}

\clearpage

\subsection{Ablations}
\label{sec:app:ablations}

Apart from model size and latent shapes several other design decisions strongly impacted model performance, in particularly for image generation. These include the normalization of the functaset, the guidance scale for classifier-free guidance, and the sampling of timesteps for diffusion training. Here we present ablations for these choices and evaluate how they affect the FID score for diffusion modeling on spatial functa for ImageNet-1k $256\times 256$.

\subsubsection{Normalization of the functaset prior to generative modeling}
\label{sec:app:ablations:normalization}
\begin{figure}[ht]
    \centering
    \includegraphics[height=.85\textheight]{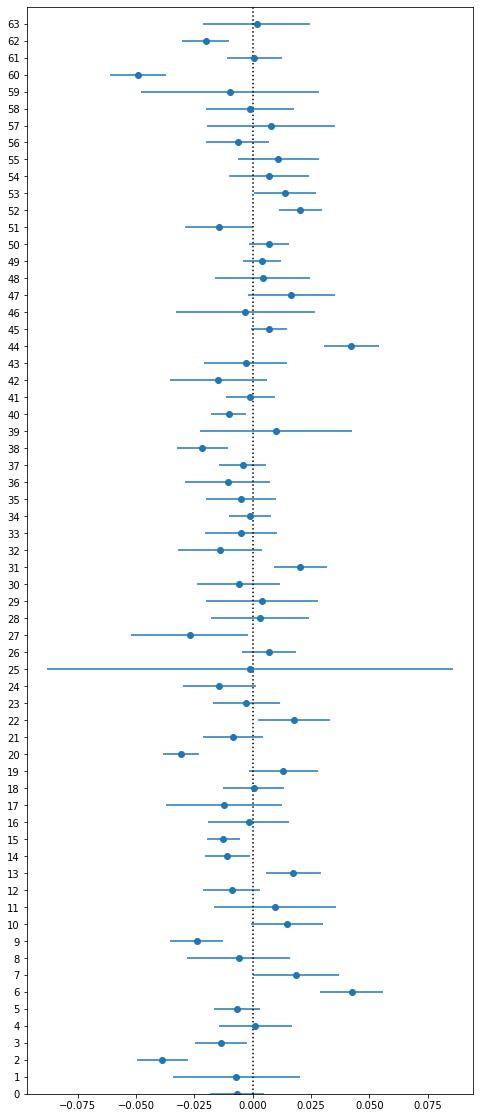}
    \caption{Mean and standard deviation of the $64$ latent feature dimensions of a functaset with spatial latents $32\times 32 \times 64$  for ImageNet-1k $256\times 256$. We compute the mean and standard deviation across spatial locations as well as examples.}
    \label{fig:imagenet_functaset_ranges}
\end{figure}

First, we investigate the normalization of the functaset before using them for downstream tasks. A priori the spatial latents $\vz \in \mathbb{R}^{s\times s\times c}$ are unconstrained, however we find that they are confined to small values, most likely due to the 3-gradient-step meta-learning procedure. In \cref{fig:imagenet_functaset_ranges} we display the means and standard deviations of the latent feature dimensions of a spatial functaset with $s=32, c=64$ on ImageNet-1k ($256\times 256$). In the figure we average over the spatial dimensions as well as over examples.

In general, we find that the distribution of the meta-learned functaset is non-Gaussian and can have heavy tails, hence we choose to standardize the mean and standard deviation of the latents to simplify the generative modeling task of the diffusion model. We find that the choice of normalization has a strong effect on the final performance. We standardize the latents as
\begin{equation}
	\widetilde{\vz} = \frac{\vz - \mu}{\gamma \cdot \sigma}
\end{equation}
where subtraction and division are element-wise and $\mu$ and $\sigma$ are means and standard deviations across the training set. In addition, $\gamma$ denotes an additional (global) scale factor that we treat as a tunable hyperparameter. 

We consider three different choices how to compute these means and standard deviations:
\begin{enumerate}
    \item scalar-valued $\mu, \sigma \in \mathbb{R}$: compute means and standard deviations over all latent dimensions $s\times s \times c$; each latent dimension is centred and scaled in the same way.
    \item vector-valued $\mu, \sigma \in \mathbb{R}^c$: compute means and standard deviations over the spatial dimensions $s \times s$, but each of the $c$ feature dimensions is centred and scaled independently of each other.
    \item array-valued $\mu, \sigma \in \mathbb{R}^{s\times s \times c}$: compute means and standard deviations only across examples, and each of the latent dimensions $s\times s \times c$ is treated independently of each other.
\end{enumerate}

\begin{figure}[ht]
    \centering
    \includestandalone[mode=image]{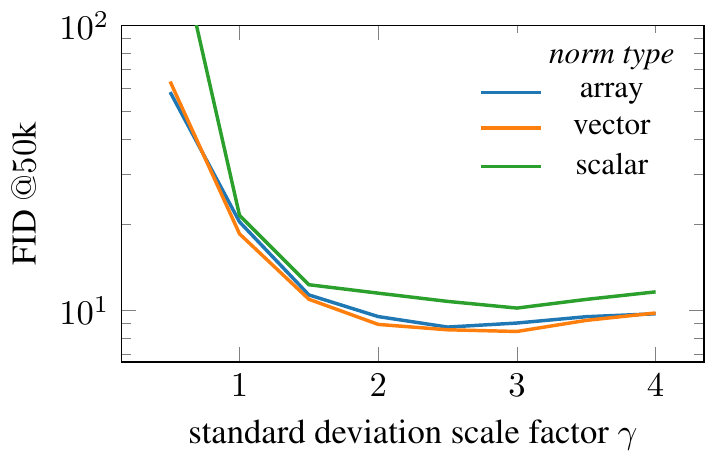}
    \caption{Ablation: Effect of different normalization types and scaling parameters used in normalizing the modulation latents for generative modeling.}
    \label{fig:imagenet_ablation_normtype_scale}
\end{figure}

We find that vector- and array-normalization yield very similar results while scalar-normalization is markedly worse. Moreover, we find that a scale factor $\gamma\approx 2.5$ yields the best results on ImageNet-256 empirically, see \cref{fig:imagenet_ablation_normtype_scale}.

\clearpage

\subsubsection{Guidance scale}

\begin{figure}[ht]
    \begin{center}%
    \includestandalone[mode=image]{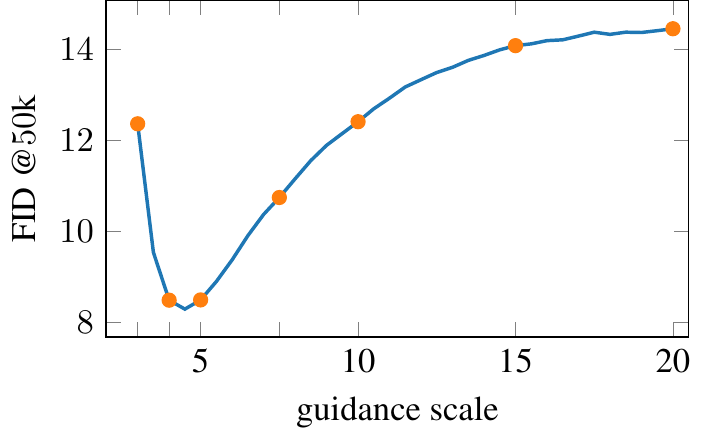}
    
    \begin{tikzpicture}
    \node[inner sep=0em, outer sep = 0em] (x) {\includegraphics[width=\textwidth]{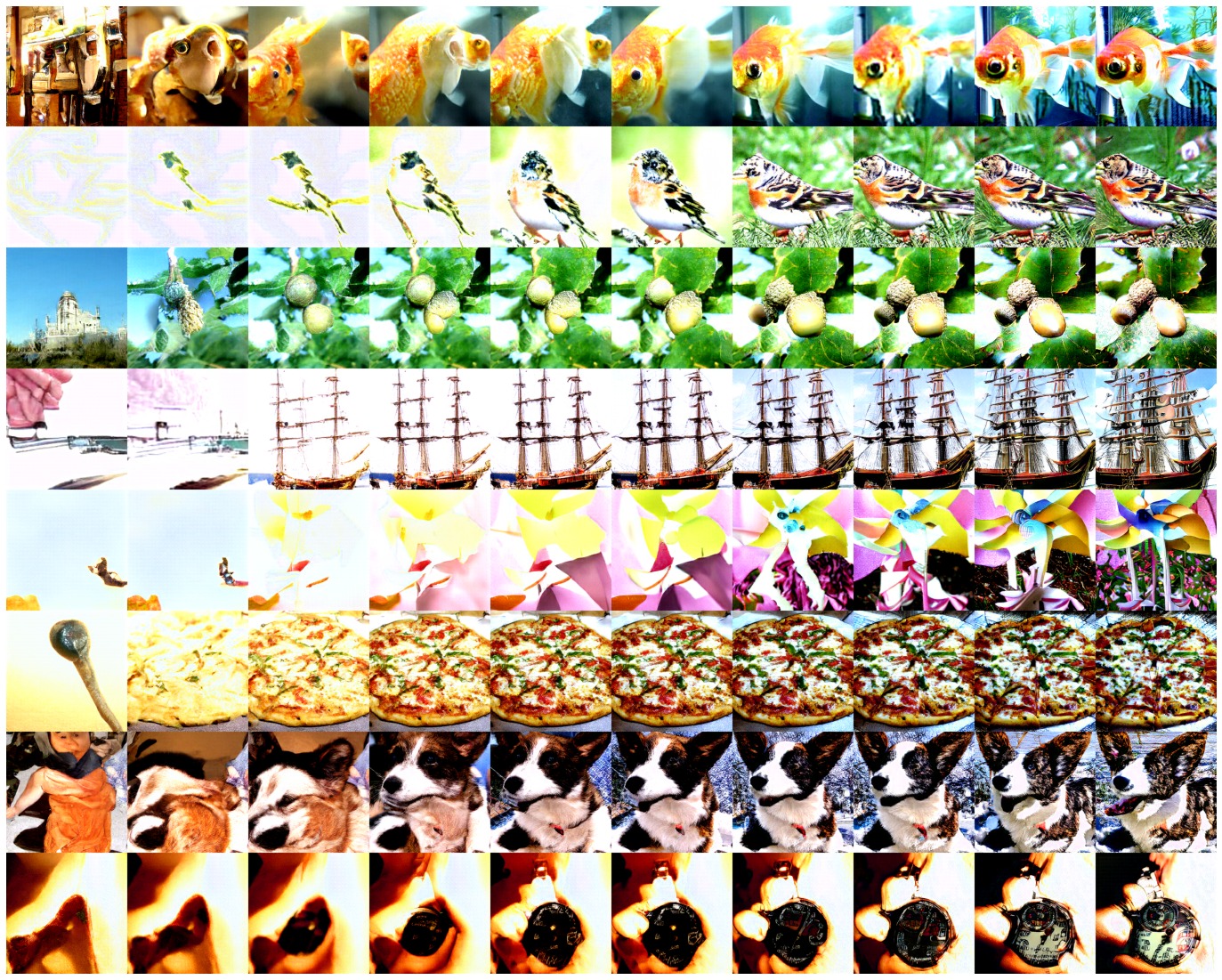}};
    \node (left) at (x.north west){};
    \node (right) at (x.north east){};
    \node[anchor=south] at ($(left)!1/20!(right)$) {\textbf{0}};
    \node[anchor=south] at ($(left)!3/20!(right)$) {\textbf{1}};
    \node[anchor=south] at ($(left)!5/20!(right)$) {\textbf{2}};
    \node[anchor=south] at ($(left)!7/20!(right)$) {\textcolor{orange}{\textbf{3}}};
    \node[anchor=south] at ($(left)!9/20!(right)$) {\textcolor{orange}{\textbf{4}}};
    \node[anchor=south] at ($(left)!11/20!(right)$) {\textcolor{orange}{\textbf{5}}};
    \node[anchor=south] at ($(left)!13/20!(right)$) {\textcolor{orange}{\textbf{7.5}}};
    \node[anchor=south] at ($(left)!15/20!(right)$) {\textcolor{orange}{\textbf{10}}};
    \node[anchor=south] at ($(left)!17/20!(right)$) {\textcolor{orange}{\textbf{15}}};
    \node[anchor=south] at ($(left)!19/20!(right)$) {\textcolor{orange}{\textbf{20}}};
    \end{tikzpicture}
    \end{center}
    \caption{Class-conditional classifier-free DDPM sampling from a diffusion model at different guidance scales ranging from $0$ (unconditional samples) to $20$. \\
    \textit{Top:} FID values using $50$k samples; \textit{bottom:} uncurated samples at different guidance scales. The orange dots in the upper plot correspond to the orange colored guidance scales $\geq 3$.}
    \label{fig:imagenet_ablation_guidance_scale_ablation}
\end{figure}
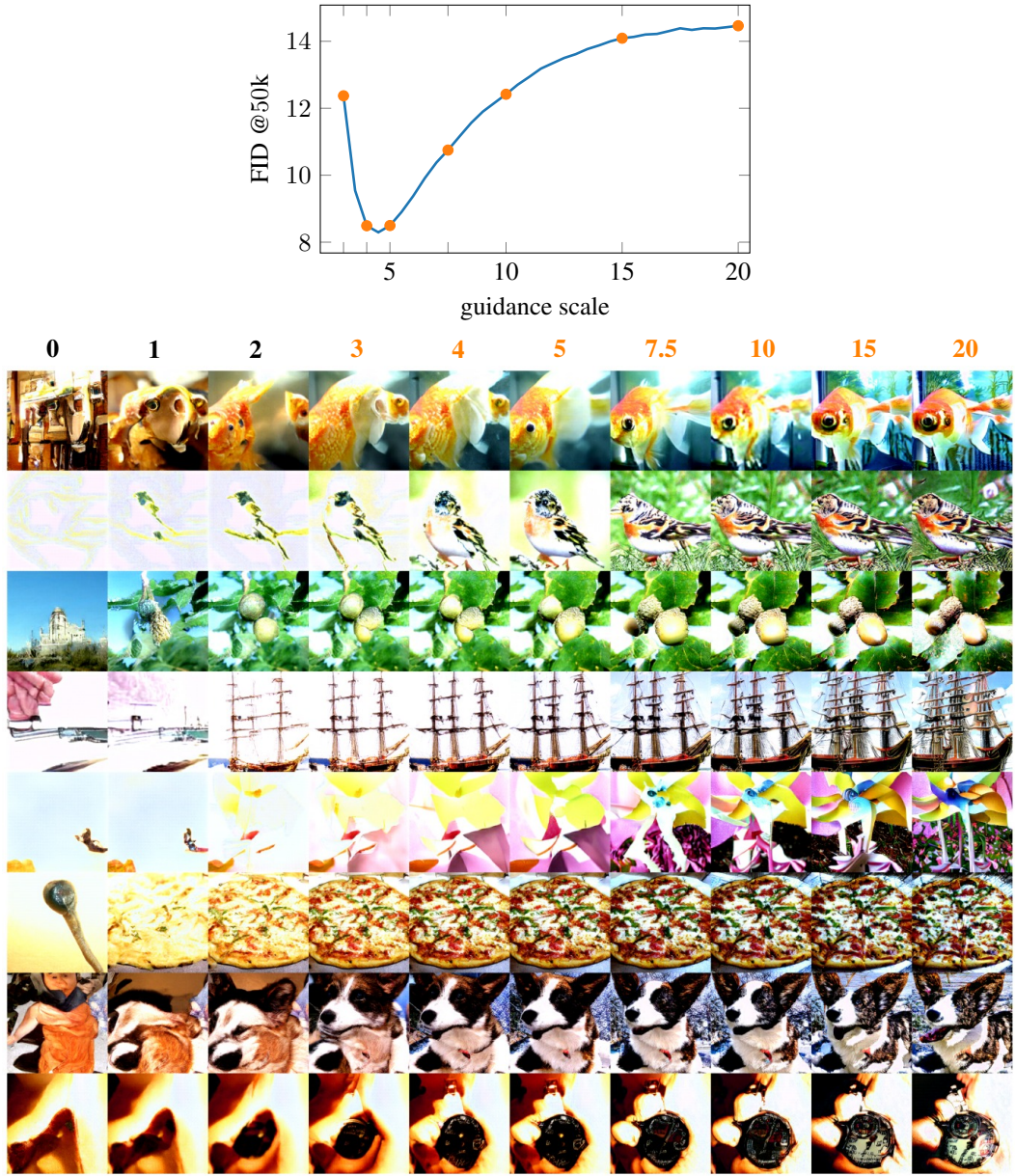

When sampling diffusion samples, we use classifier-free guidance \citep{ho2022classifier} to improve the image quality. 
In this ablation, we sweep the guidance scale for ImageNet-1k $256\times 256$ and find that the optimal guidance scale is between $4$ and $5$, see \cref{fig:imagenet_ablation_guidance_scale_ablation} \textit{(top)}. This range is larger than the range typically found for diffusion modeling in pixel space \citep{ho2022classifier} or latent space \citep{rombach2022high}. In all our experiments in the main paper we use classifier-free guidance with a guidance scale of $5$. In \cref{fig:imagenet_ablation_guidance_scale_ablation} \textit{(bottom)} we show uncurated samples for different classes and for different values of the guidance scale. Too small guidance scales lead to unclear samples whereas too large guidance scales lead to patch boundary artifacts or exaggerated contrast.

\subsubsection{Ratio of time step sampling during diffusion training}

\begin{figure}[h]
    \centering
    \includestandalone[mode=image]{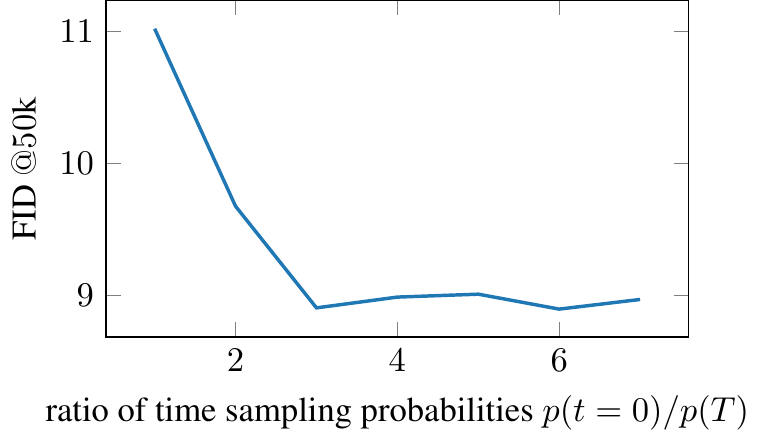}
    \caption{Ablation: Effect of sampling diffusion time steps with varying frequencies during training. A ratio of $1$ corresponds to uniform sampling.}
    \label{fig:imagenet_ablation_timestep_sampling_training}
\end{figure}

For generative diffusion modeling we use the discrete-time formulation of diffusion models (DDPM) \citep{Ho2020}. While their simple training objective suggests to sample time-steps uniformly during training, $t\sim\mathrm{Uniform}(0, T)$, we found that sampling earlier time steps more frequently lead to improved performance. We sample timesteps according to a distribution that linearly interpolates between a maximal probability value for timesteps $t\rightarrow 0$ and a minimal value for timesteps $t\rightarrow T$. In \cref{fig:imagenet_ablation_timestep_sampling_training} we vary the ratio between these two extremes ranging from $1$ (uniform) to $7$. We found that using a ratio of ${}\geq 3$ gave the best results.

\subsubsection{Quantization}
\begin{figure}[h]
    \centering
	\begin{tikzpicture}
	\tikzstyle{every node}=[font=\small]
        \node[inner sep=0em, outer sep = 0em] (x) {\includegraphics[width=0.7\textwidth]{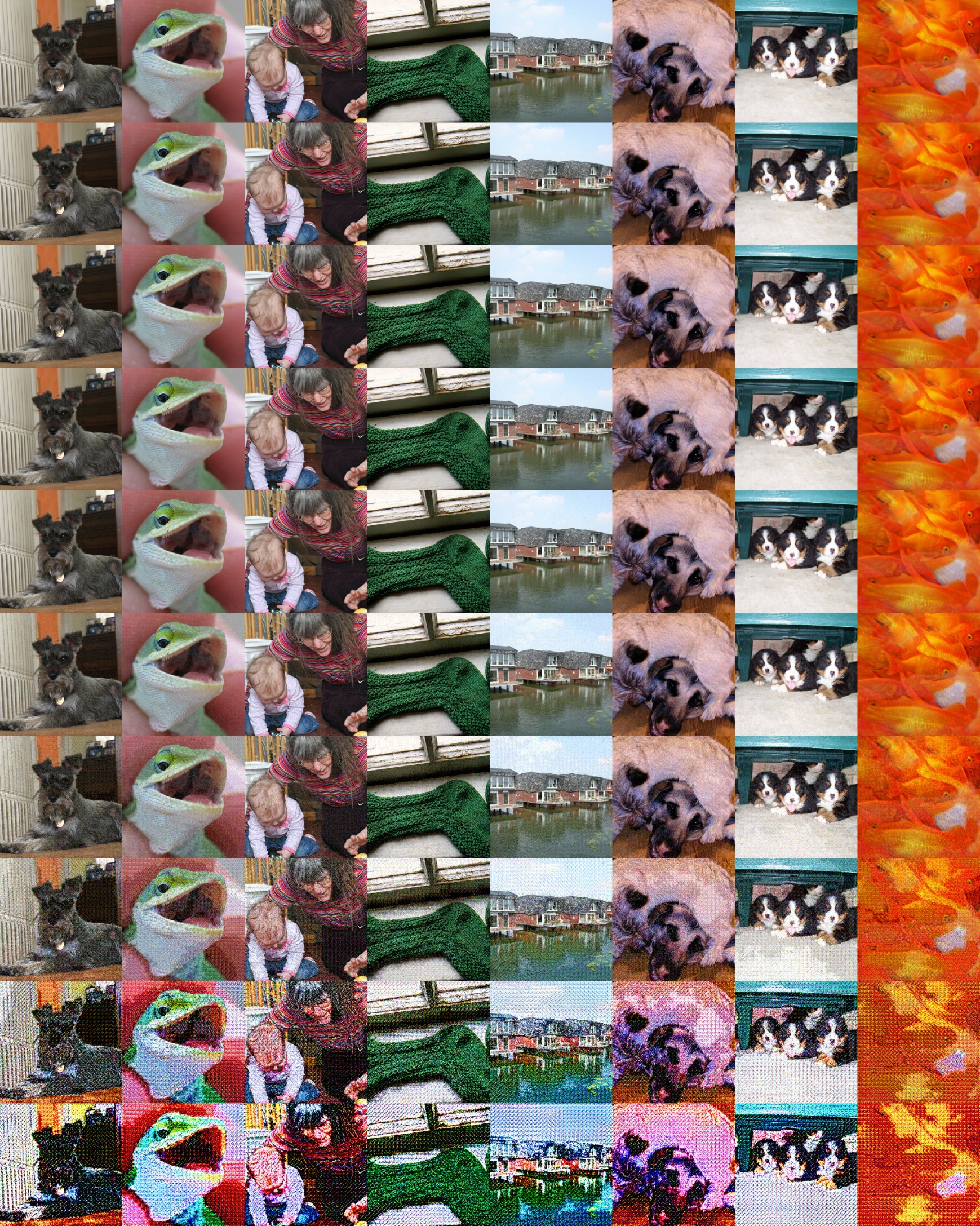}};
        \node (top) at (x.north west){};
        \node (bottom) at (x.south west){};
        \node[anchor=east] at ($(top)!1/20!(bottom)$) {Original};
        \node[anchor=east] at ($(top)!3/20!(bottom)$) {unquantized float32: \textbf{36.6 dB}};
        \node[anchor=east] at ($(top)!5/20!(bottom)$) {8 bit: \textbf{36.4 dB}};
        \node[anchor=east] at ($(top)!7/20!(bottom)$) {7 bit: \textbf{36.1 dB}};
        \node[anchor=east] at ($(top)!9/20!(bottom)$) {6 bit: \textbf{35.0 dB}};
        \node[anchor=east] at ($(top)!11/20!(bottom)$) {5 bit: \textbf{32.2 dB}};
        \node[anchor=east] at ($(top)!13/20!(bottom)$) {4 bit: \textbf{27.2 dB}};
        \node[anchor=east] at ($(top)!15/20!(bottom)$) {3 bit: \textbf{20.9 dB}};
        \node[anchor=east] at ($(top)!17/20!(bottom)$) {2 bit: \textbf{14.4 dB}};
        \node[anchor=east] at ($(top)!19/20!(bottom)$) {1 bit: \textbf{11.7 dB}};
    \end{tikzpicture}
    \caption{PSNR of different quantization levels for spatial functa shape $32\times32\times64$ on ImageNet $256 \times 256$ reconstructions. Quantization is per latent dimension, binning each into $2^{\text{\#bits}}$ uniform bins. Best viewed zoomed in.}
    \label{fig:quantization_ablation}
\end{figure}
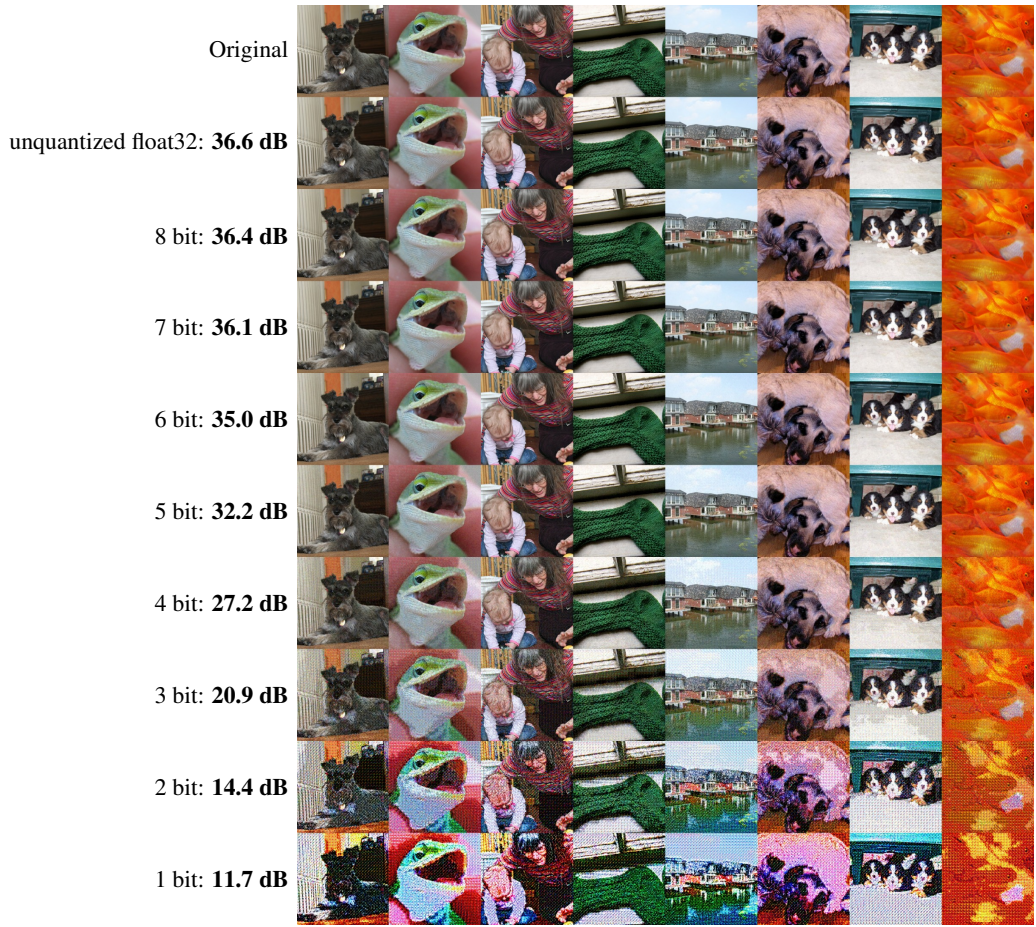

We show that spatial functa are highly compressible, making them amenable to neural-field based compression methods described in e.g.\ \cite{dupont2021coin, dupont2022coin++, strumpler2022implicit, schwarz2022meta}. In \cref{fig:quantization_ablation}, we show that na{\"i}ve uniform quantization of each 32 bit latent dimension of spatial functa leads to hardly any decrease in reconstruction quality down to 6 bits per dimension. In \cref{tab:imagenet_quantization_ablation} we further show FID scores of class-conditional diffusion models trained on each of the quantized modulations, showing that there is hardly a loss in perceptual quality of samples down to 6 bits.

\begin{table}[ht]
\centering
\setlength{\arrayrulewidth}{0.3mm}
\begin{tabular}{c | CCCC}
\toprule
\textbf{Num bits per latent dimension} & 32~ (\text{unquantized}) & 8 & 7 & 6 \\
\midrule
\textbf{FID}@50k (cond) & 8.7 \pm 0.2 & 9.0 \pm 0.2 & 8.8 \pm 0.2 & 9.2 \pm 0.1 \\
\bottomrule
\end{tabular}
\caption{ImageNet $256\times256$ diffusion ablation on different levels of uniformly quantized spatial functa. We show means and standard deviations of FID scores over 3 random seeds.}
\label{tab:imagenet_quantization_ablation}
\end{table}

\clearpage
\subsection{Perturbation analysis of spatial functasets}

Here, we provide a brief perturbation analysis on spatial functa, similar to the analysis performed by \citet{functa} as well as that presented in \cref{fig:cifar10_functa_perturbation}.

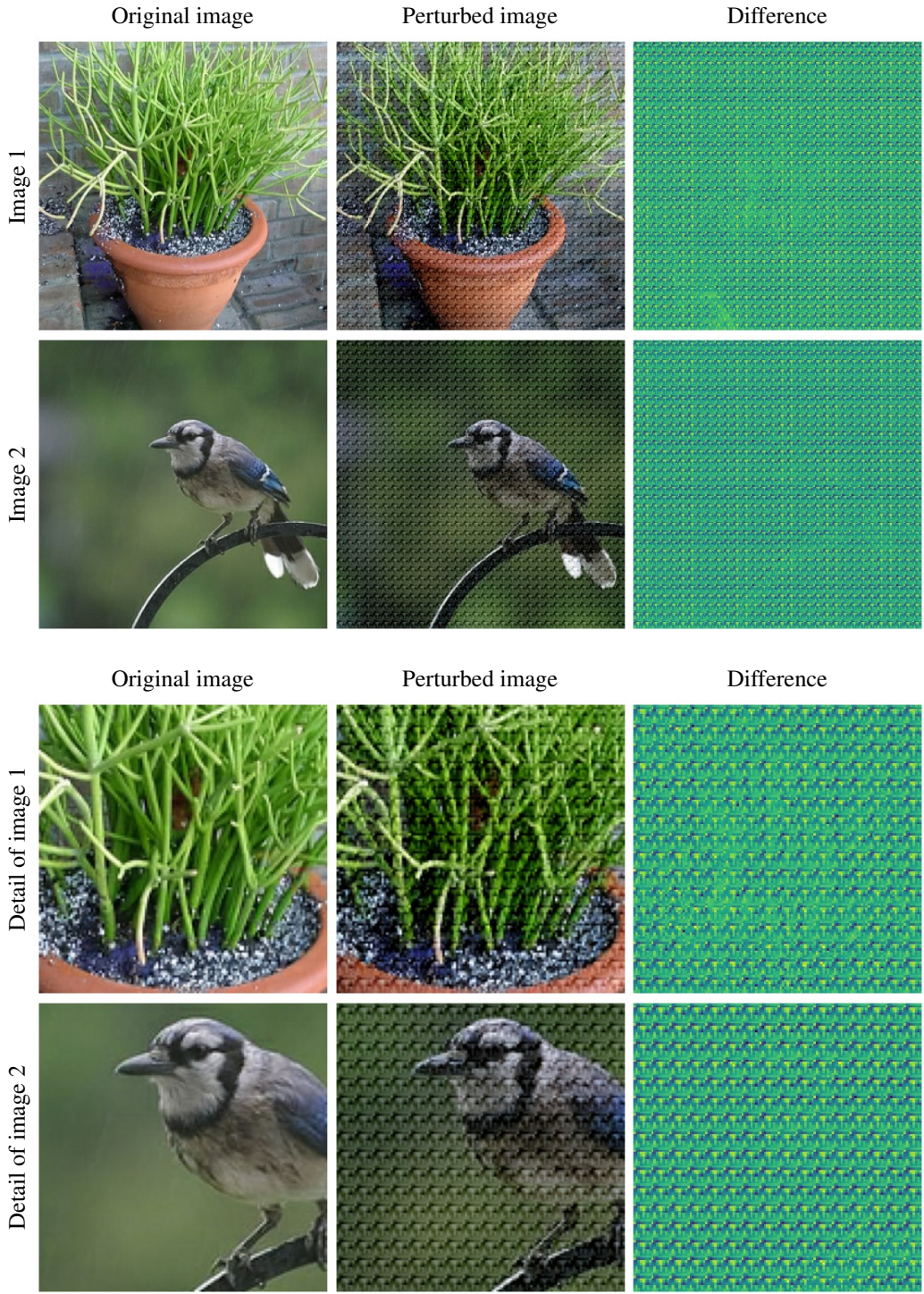
\begin{figure}
    \centering
    \begin{tikzpicture}
    \node[inner sep=0em, outer sep = 0em] (x) {\includegraphics[width=0.95\textwidth]{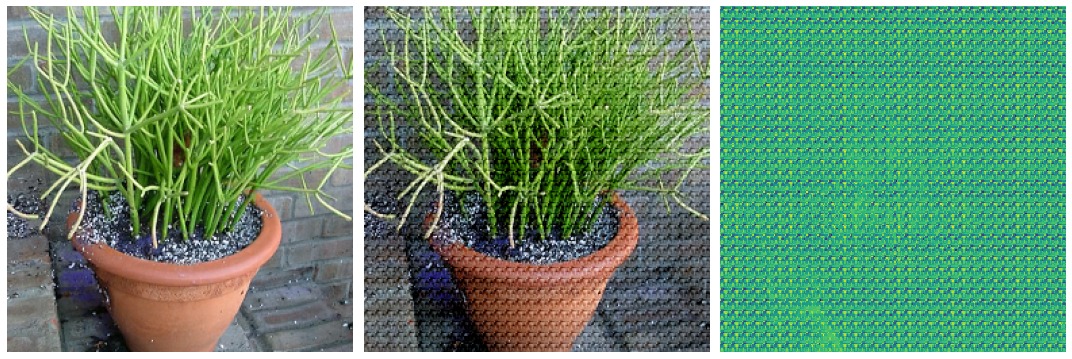}};
    \node (left) at (x.north west){};
    \node (right) at (x.north east){};
    \node[anchor=south] at ($(left)!1/6!(right)$) {Original image};
    \node[anchor=south] at ($(left)!3/6!(right)$) {Perturbed image};
    \node[anchor=south] at ($(left)!5/6!(right)$) {Difference\vphantom{g}};
    
    \node[below= 0 of x, inner sep=0em, outer sep = 0em] (x2) {\includegraphics[width=0.95\textwidth]{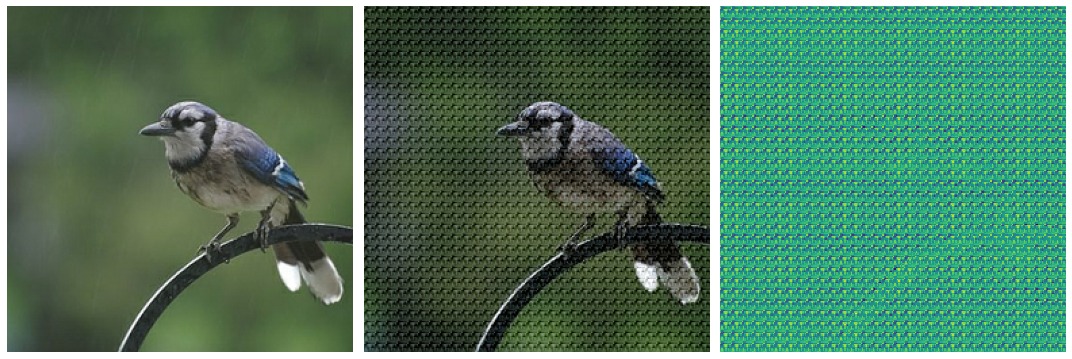}};
    
    \node[left = 0.5 of x.west,  anchor=north, rotate=90]{Image 1};
    \node[left = 0.5 of x2.west,  anchor=north, rotate=90]{Image 2};
    \end{tikzpicture}    
    
    \
    
    \begin{tikzpicture}
    \node[inner sep=0em, outer sep = 0em] (x) {\includegraphics[width=0.95\textwidth]{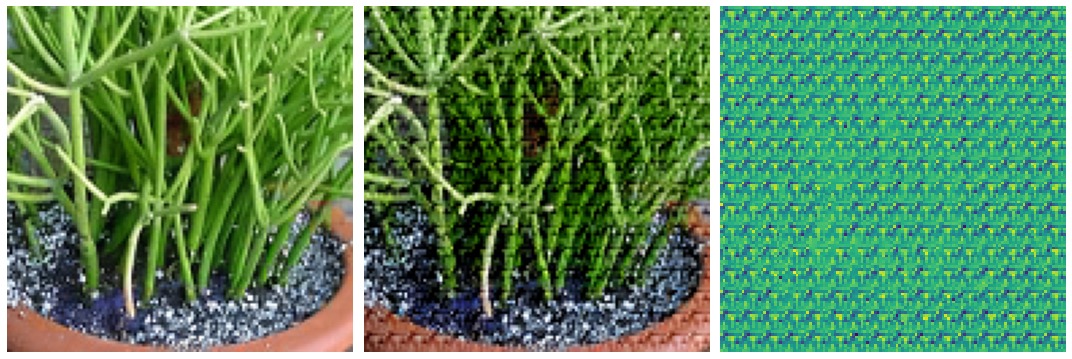}};
    \node (left) at (x.north west){};
    \node (right) at (x.north east){};
    \node[anchor=south] at ($(left)!1/6!(right)$) {Original image};
    \node[anchor=south] at ($(left)!3/6!(right)$) {Perturbed image};
    \node[anchor=south] at ($(left)!5/6!(right)$) {Difference\vphantom{g}};
    \node[below= 0 of x, inner sep=0em, outer sep = 0em] (x2) {\includegraphics[width=0.95\textwidth]{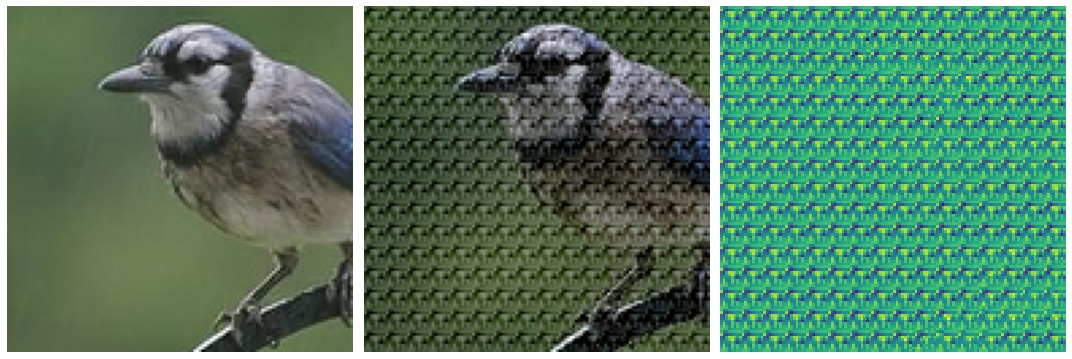}};
    
    \node[left = 0.5 of x.west,  anchor=north, rotate=90]{Detail of image 1};
    \node[left = 0.5 of x2.west,  anchor=north, rotate=90]{Detail of image 2};
    \end{tikzpicture} 
    \caption{Example of a spatial functa perturbation. Here, we perturb the $26$th feature dimension of every $s\times s$ spatial latent by adding the same constant value to each while keeping all other feature dimensions unmodified. \textit{top:} full $256\times 256$ image; \textit{bottom:} zoomed in detail}
    \label{fig:imagenet_perturbation_spatial_latent_sample}
\end{figure}

In our experiments, we perturbed individual feature dimensions at each spatial location of the $s\times s$ spatial functa latent by the same constant value, see \cref{fig:imagenet_perturbation_spatial_latent_sample} for an example where we perturb the $26$th (among the $c=64$) feature dimension at each spatial location for two images from ImageNet-1k ($256\times 256$). 
We observed that the difference between the original and perturbed image has a repeating pattern of size $d/s$, where $d$ is the image size in pixels and $s$ is the latent size. We found that the pattern is stable across patches of the same image as well as between images; thus, each latent feature dimension explains the same local feature in each patch. Note that the patterns are somewhat modulated in darker or brighter regions where the pixel values saturate. As the strength of the perturbation varies, so does the strength of the pattern. 
To some degree we can think of each feature dimension as coefficients of basis functions in the pixel space, though we only inspect perturbations of individual dimensions.

\begin{figure}
    \centering
    \includegraphics[width=\textwidth]{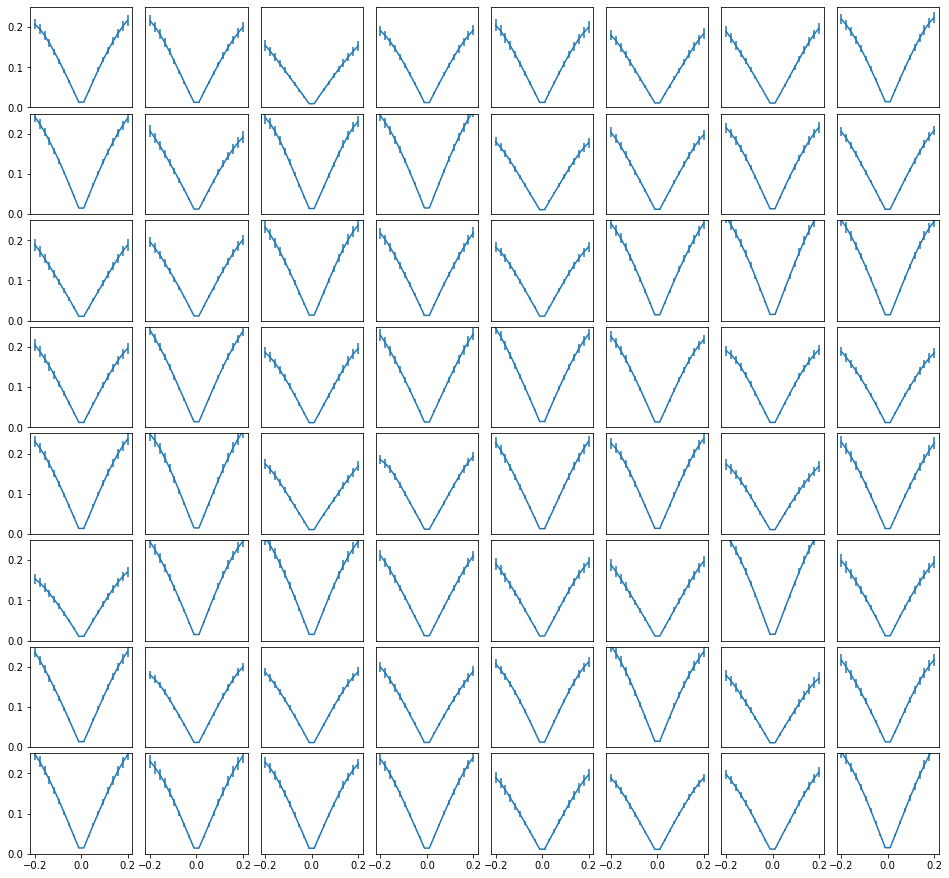}
    \caption{Perturbation analysis when independently perturbing each of the $64$ feature dimension of every spatial latent by adding the same constant value to each while keeping all other feature dimensions unmodified. Each plot corresponds to one of the $64$ feature dimensions.\\
    In each plot, the $x$-axis denotes the perturbation strength and the $y$-axis corresponds to the RMSE between the perturbed images and the original image.}
    \label{fig:imagenet_perturbation_spatial_latent}
\end{figure}

We further analyze how the strength of the pattern changes as we vary the strength of the perturbation. In \cref{fig:imagenet_functaset_ranges} we saw that the typical range of the latent modulations is $\leq 0.1$; we therefore explore perturbations with a strength in the range $[-0.2, 0.2]$. Like in \cref{fig:imagenet_perturbation_spatial_latent_sample} we perturb each of the $c=64$ feature dimension separately but at each of the $s\times s = 32\times 32$ spatial dimensions simultaneously. In \cref{fig:imagenet_perturbation_spatial_latent} we show the average RMSE for each of the $c=64$ feature dimensions computed over $8 \times 8$ patches (corresponding to the repeating pattern in \cref{fig:imagenet_perturbation_spatial_latent_sample}) and images. Interestingly, we observe that for all of the $c=64$ feature dimensions the RMSE ($y$-axes) increases approximately linearly with the perturbation strength ($x$-axes). These preliminary results hint at the possibility that the latent feature dimensions encode coefficients of a set of linear basis functions, though further work in this direction is needed for verification.

\end{document}